\documentclass[11pt]{article}

\usepackage[final]{acl}

\usepackage{times}
\usepackage{latexsym}
\usepackage{graphicx}
\usepackage{stfloats}
\usepackage{tabularx}
\usepackage{booktabs}
\usepackage{hyperref}
\usepackage{multicol}
\usepackage{fontawesome5}
\usepackage[T1]{fontenc}
\usepackage[utf8]{inputenc}

\usepackage{microtype}
\usepackage{tcolorbox}
\usepackage{xcolor}

\definecolor{TiffanyBlue}{RGB}{129,216,207}
\tcbuselibrary{listings,theorems}
\newtcolorbox{mybox}{colback=white!5!white,colframe=black!75!black, left=.05in, right=.05in}
\newtcbtheorem[]{exmp}{Example}%
{colback=TiffanyBlue!5,colframe=TiffanyBlue!80,fonttitle=\bfseries, left=.08in, right=.08in,bottom=.05in, top=.05in}{exmp}
\usepackage{inconsolata}

\title{EternalMath: A Living Benchmark of Frontier Mathematics that Evolves with Human Discovery}

\author{
  \textbf{Jicheng Ma}\textsuperscript{1}, 
  \textbf{Guohua Wang}\textsuperscript{2,}\thanks{~Corresponding author.}, 
  \textbf{Xinhua Feng}\textsuperscript{2}, 
  \textbf{Yiming Liu}\textsuperscript{2},
  \textbf{Richeng Xuan}\textsuperscript{2},
  \textbf{Zhichao Hu}\textsuperscript{2,}\thanks{~Project leader.},
  \textbf{Yuhong Liu}\textsuperscript{2} \\
  \textsuperscript{1}School of Mathematics, Renmin University of China, Beijing, China \\
  \textsuperscript{2}Tencent, China \\
  \texttt{guohuawang@tencent.com}
}

\begin{document}
\maketitle
\begin{figure*}[b]
  \centering
  \begin{minipage}{0.48\textwidth}
    \centering
    \includegraphics[width=\linewidth]{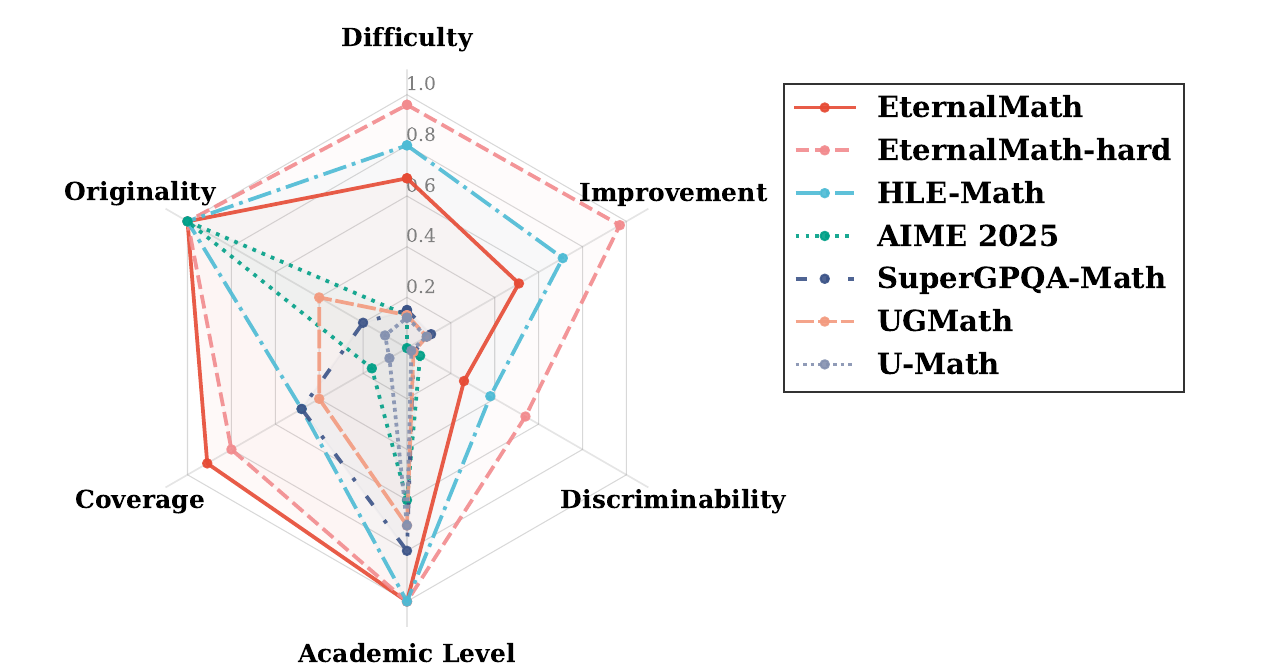}
    \caption{Benchmark discriminative power.(App. \ref{app:radar})}
    \label{fig:radar}
  \end{minipage}
  \hfill 
  \begin{minipage}{0.48\textwidth}
    \centering
    \includegraphics[width=\linewidth]{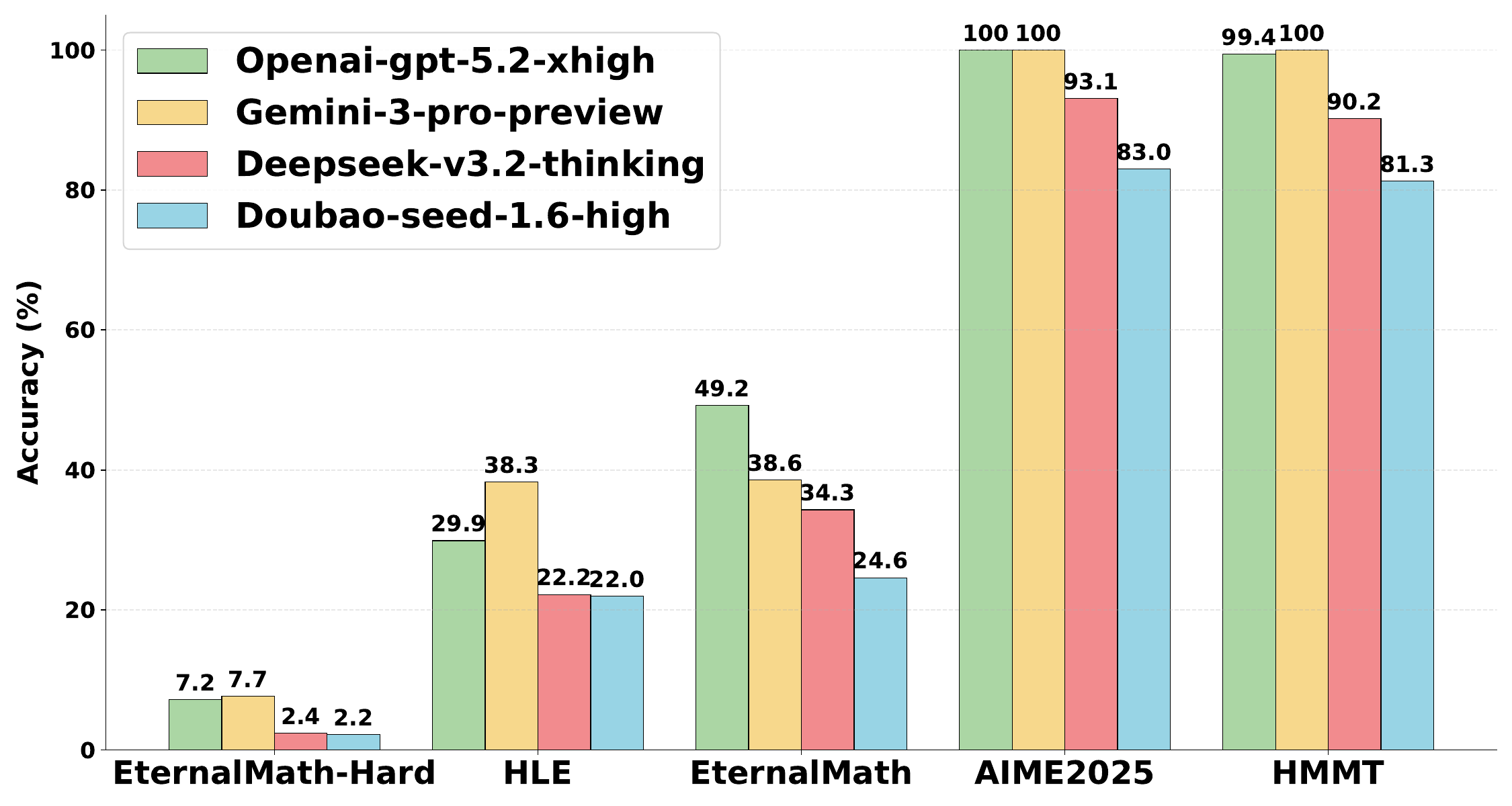}
    \caption{SOTA model performance.}
    \label{fig:performance}
  \end{minipage}
  %\caption{Benchmark discriminative power (App. \ref{app:radar}). Right: SOTA model performance comparison.} 
  % \label{fig:my_combined_figure}
\end{figure*}
\begin{abstract}
Current evaluations of mathematical reasoning in large language models (LLMs) are dominated by static benchmarks, either derived from competition-style problems or curated through costly expert effort, resulting in limited coverage of research-level mathematics and rapid performance saturation. We propose a fully automated, theorem-grounded pipeline for evaluating frontier constructive and quantitative mathematical reasoning, which directly transforms recent peer-reviewed mathematical literature into executable and verifiable reasoning tasks. The pipeline identifies constructive or quantitative results, instantiates them into parameterized problem templates, and generates deterministic solutions through execution-based verification, enabling scalable, reproducible, and continuously updatable evaluation without reliance on large-scale expert authoring. By design, this approach supports temporal extensibility and domain-specific customization across mathematical subfields. Applying this pipeline yields \textbf{EternalMath}, an evolving evaluation suite derived from contemporary research papers. Experiments with state-of-the-art LLMs reveal substantial performance gaps, indicating that mathematical reasoning at the research frontier remains far from saturated and underscoring the need for evaluation methodologies that evolve in step with human mathematical discovery.
\end{abstract}

\section{Introduction}

Large Language Models (LLMs) have recently demonstrated remarkable progress in mathematical reasoning, evolving from fluent text generators into systems capable of solving increasingly complex formal problems. These advances are attributed to architectural refinements \cite{mcleish2024transformers}, advanced reinforcement learning paradigms \cite{wang2024math,shao2025deepseekmath}, and effective prompting strategies \cite{wei2022chain}. Consequently, frontier models now achieve near-saturated performance on foundational benchmarks such as GSM8K \cite{cobbe2021GSM8k} and MATH \cite{hendrycks2021measuring}, leading to a decline in their discriminative power. Furthermore, as these datasets are derived from public competitions and textbooks, they are increasingly susceptible to data contamination \cite{balunovic2025matharena}, making it difficult to distinguish genuine reasoning from memorized patterns.

To address these limitations, recent initiatives such as FrontierMath \cite{glazer2024frontiermath} and Humanity’s Last Exam (HLE) \cite{phan2025humanitysexam} have shifted toward expert-curated, high-difficulty problems. While effective in mitigating direct leakage, this expert-centric paradigm faces significant sustainability challenges. First, the heavy reliance on human experts—with HLE requiring contributions from nearly 1,000 specialists—makes these benchmarks expensive to produce and slow to iterate. Second, such problems are often designed as isolated, artificially abstracted puzzles that lack the structural depth of real-world mathematical research. Most importantly, once published, these static datasets become immediate targets for future training data ingestion, leading to a recurring cycle of benchmark obsolescence.

In this work, we propose a shift from static dataset curation to an automated, theorem-grounded generation methodology. We introduce a multi-agent pipeline designed to transform recent peer-reviewed mathematical research into verifiable and parameterized reasoning tasks. Unlike prior automated methods that rely on unstructured LLM synthesis \cite{yu2023metamath, tang2024mathscale}, our approach anchors the generation process in the constructive results of published literature. By extracting quantitative kernels from research papers and implementing them as executable Python-based instantiations, our pipeline ensures both mathematical rigor and objective verifiability.

This methodology offers three distinct advantages over traditional benchmarks. First, it enables temporal scalability: the pipeline can continuously ingest newly published papers from repositories like arXiv or top mathematics journals, creating a dynamically evolving evaluation set that stays ahead of model training cutoffs. Second, it provides intrinsic verification: by converting abstract theorems into executable code, the pipeline generates deterministic ground truths, eliminating the need for costly human grading or unreliable LLM-based evaluation. Third, it allows for domain-specific customization: the framework can be restricted to specific subfields (e.g., Algebraic Geometry or Partial Differential Equations), enabling targeted assessment of specialized mathematical capabilities.

Leveraging this pipeline, we present \textbf{EternalMath}, a benchmark consisting of research-level problems derived from the latest mathematical discoveries. Our evaluation of frontier LLMs on \textbf{EternalMath} reveals a substantial performance gap compared to traditional benchmarks. Even state-of-the-art models struggle to maintain logical consistency across the multi-stage derivations required by our theorem-grounded tasks, highlighting persistent limitations in handling real-world research mathematics.

In summary, our contributions are threefold:
\begin{itemize}
  \item We propose an executable, theorem-grounded benchmark construction pipeline that transforms recent peer-reviewed mathematical results into parameterized and verifiable problem instances, enabling scalable and low-cost generation of research-level evaluation data without large-scale expert authoring.
  \item We introduce \textbf{EternalMath}, an instantiation of this pipeline that yields a dynamically extensible mathematical reasoning benchmark grounded in real-world research problems, with high difficulty, strong discriminative power and robust resistance to data contamination.\footnote{To ensure full reproducibility, the complete dataset, parameterized meta-templates, and executable Python scripts are publicly available at \href{https://anonymous.4open.science/r/anonymous-403B}{anonymous link}.}
  \item We evaluate state-of-the-art LLMs on \textbf{EternalMath} and demonstrate that frontier-level constructive and quantitative mathematical reasoning remains far from saturated, revealing persistent limitations of LLMs on real-world research-level mathematical tasks.
\end{itemize}

\section{Related Work}
Mathematical reasoning evaluation for Large Language Models (LLMs) has evolved rapidly, yet existing paradigms face inherent limitations in assessing real-world research capabilities.

\paragraph{Static and Expert-Curated Benchmarks} 
Existing static datasets, ranging from foundational sets (GSM8K~\cite{cobbe2021GSM8k}, GAOKAO~\cite{zhang2023evaluating}) to complex competition-level problems (MATH~\cite{hendrycks2021measuring}, Omni-MATH~\cite{gao2024omni}, PutnamBench~\cite{tsoukalas2024putnambench}), suffer from rapid performance saturation and are highly susceptible to data contamination~\cite{balunovic2025matharena}. To maintain rigor, recent initiatives like FrontierMath~\cite{glazer2024frontiermath} and Humanity’s Last Exam (HLE)~\cite{phan2025humanitysexam} rely on expert-authored tasks. While providing high-quality signals, expert curation is inherently expensive, slow to update, and difficult to scale. Furthermore, these artificially abstracted puzzles often differ from the structural depth of organic mathematical research.

\paragraph{Automated Generation and Verifiable Evaluation} 
To address saturation and contamination, automated and dynamic evaluations have emerged. Problem-driven (MetaMath~\cite{yu2023metamath}, DART-Math~\cite{tong2024dart}) and knowledge-driven methods (MathScale~\cite{tang2024mathscale}, SAND-Math~\cite{manem2025sand}, STORM-BORN~\cite{liu2025storm}) offer scalability but often struggle to bridge the gap between synthetic breadth and research-level depth. Orthogonally, formal proof benchmarks (MiniF2F~\cite{zheng2021minif2f}, Lean-Workbook~\cite{ying2024lean}) provide strict verifiability but require fully formalized corpora, making them less suitable for evaluating natural language reasoning in open research. Recent efforts like RealMath~\cite{zhang2025realmath} extracts verifiable QA pairs from live arXiv papers but generates static questions without dynamic parameterization. LemmaBench~\cite{peyronnet2026lemmabench} evaluates models on live research proofs but relies on potentially biased LLM-as-a-judge evaluation. Conversely, VAR-MATH~\cite{yao2025var} utilizes programmatic parameterization for multi-instance generation but restricts its source to static, historical corpora.

\textbf{Our Scope:} \textbf{EternalMath} uniquely intersects these paradigms. By strictly focusing on \textbf{constructive and computable problems}, we extract quantitative kernels from the live research frontier, apply programmatic parameterization to generate diverse variants, and evaluate them via deterministic code execution. This framework simultaneously overcomes the static nature of traditional datasets, the high cost of expert curation, and the reliability issues of LLM judges, offering a sustainable, contamination-resistant evaluation.

\begin{figure*}[hbtp]
  \includegraphics[width=\linewidth]{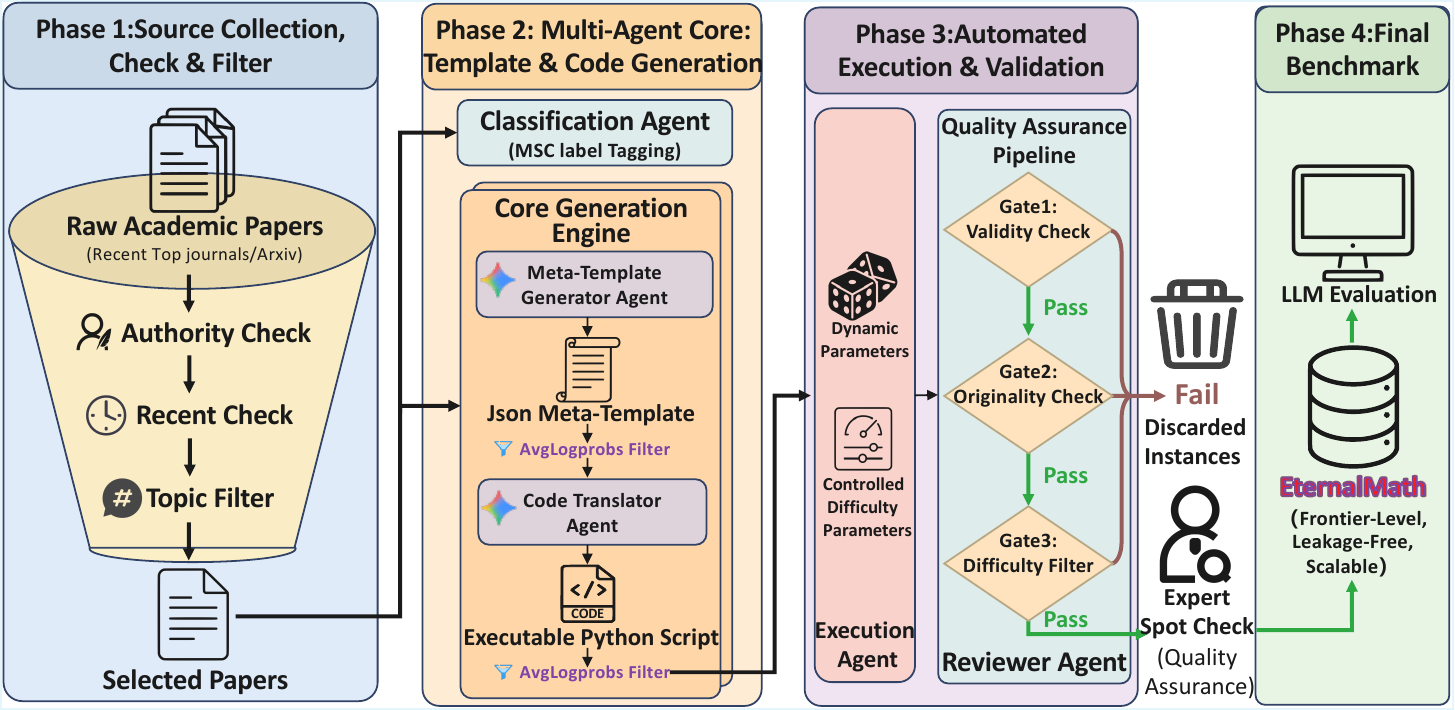}
  \caption{Overview of the \textbf{EternalMath} construction pipeline. The process consists of four primary stages: (1) \textbf{Paper Filtering}, which selects high-quality, computable research papers from top-tier venues; (2) \textbf{Multi-agent Collaboration}, where specialized agents transform theorems into parameterized meta-templates and executable Python scripts; (3) \textbf{Execution \& Verification}, utilizing symbolic computation to ensure deterministic and correct solution generation; and (4) \textbf{Validation \& Quality Control}, involving model-based difficulty stratification and human auditing to ensure benchmark rigor and contamination resistance.}
  \label{fig:pipeline}
\end{figure*}

\section{Overall Pipeline}
Figure~\ref{fig:pipeline} illustrates the \textbf{EternalMath} construction pipeline. To enable scalable, reproducible, and contamination-resistant evaluation, we ground problem generation directly in recent peer-reviewed literature rather than relying on manual expert authoring. This modular framework preserves a tight coupling between abstract theorems, concrete problem instances, and executable verification.

\subsection{Paper Filtering}
We compile a corpus of high-quality research papers using a strict filtering procedure based on three criteria:

\paragraph{Recency}
We prioritize papers published within the past two years. This directly targets the mathematical research frontier and structurally mitigates the risk of test set contamination in existing LLMs.

\paragraph{Authority}
To ensure mathematical rigor and reliability, all selected sources originate from top-tier peer-reviewed mathematics journals and vetted arXiv preprints, strictly excluding insufficiently validated results.

\paragraph{Computability}
We prioritize theorems with constructive or quantitative formulations that admit explicit input–output relationships (e.g., computing an object $C$ given conditions $B$). By excluding purely qualitative existence proofs, we ensure problems can be programmatically instantiated and verified within a closed-loop pipeline. Importantly, such constructive or computable results are pervasive across diverse mathematical landscapes, extending far beyond traditionally computational subfields.

\subsection{Multi-agent collaboration}

To transform mathematical literature into high-quality, verifiable benchmark instances, we adopt a structured multi-agent framework composed of four specialized agents. This design decomposes the complex generation process into distinct and interpretable sub-tasks, enabling reliable automation while maintaining mathematical correctness.

We employ a multi-agent architecture for two reasons. First, the end-to-end task extracting research-level theorems, instantiating them into concrete problems, and verifying their solutions involves heterogeneous objectives that are difficult to satisfy within a single prompt or model invocation. Second, separating responsibilities allows each agent to operate under narrowly scoped constraints, improving stability, controllability, and reproducibility, yielding a superior result compared to a single LLM generation of Question-Answer (QA) pairs~\cite{zhang2025realmath}. This framework enables difficulty control and verification, which are not possible in single-pass generation. {The full system prompts for all agents are provided in Appendix \ref{prompt}.}
\paragraph{Classification Agent} 
The Classification Agent assigns up to three MSC2020~\cite{dunne2020mathematics} subject codes to each selected paper, prioritizing author-provided classifications when available. This step enables structured analysis and systematic filtering. Papers belonging to non-core or descriptive categories (e.g., 00 General and overarching topics; collections, 01 History and biography, and 97 Mathematics education) are excluded to ensure suitability for quantitative problem generation.

In particular, we curated and categorized approximately 400 research articles published within the past year, organized by mathematical subfields. This classification allows the pipeline to selectively draw from specific areas of interest, supporting the construction of domain-targeted benchmark variants. Table~\ref{tab:msc-distribution} reports the distribution of these papers across disciplines and their corresponding proportions.

\begin{table*}[htbp]
\centering
\small
\begin{tabularx}{\textwidth}{@{}p{5.5cm}r@{\hspace{0.75cm}}p{6cm}r@{}}
\toprule
\textbf{MSC Classification} & \textbf{Percentage} & \textbf{MSC Classification} & \textbf{Percentage} \\
\midrule
35 Partial differential equations & 11.9\% & 18 Category theory; homological algebra & 2.9\% \\
05 Combinatorics & 9.8\% & 13 Commutative algebra & 2.7\% \\
20 Group theory and generalizations & 7.0\% & 82 Statistical mechanics, structure of matter & 2.2\% \\
14 Algebraic geometry & 6.3\% & 76 Fluid mechanics & 2.2\% \\
11 Number theory & 5.1\% & 15 Linear and multilinear algebra; matrix theory & 2.0\% \\
46 Functional analysis & 4.5\% & 17 Nonassociative rings and algebras & 2.0\% \\
65 Numerical analysis & 4.3\% & 22 Topological groups, Lie groups & 1.8\% \\
81 Quantum theory & 3.9\% & 60 Probability theory and stochastic processes & 1.8\% \\
37 Dynamical systems and ergodic theory & 3.7\% & 03 Mathematical logic and foundations & 1.6\% \\
53 Differential geometry & 3.5\% & 49 Calculus of variations and optimal control & 1.6\% \\
16 Associative rings and algebras & 3.5\% & 58 Global analysis, analysis on manifolds & 1.6\% \\
34 Ordinary differential equations & 2.9\% & Other (< 1.5\% each) & 11.4\% \\
\bottomrule
\end{tabularx}
\caption{Distribution of Mathematics Subject Classification (MSC) 2020 codes by percentage in the dataset.}
\label{tab:msc-distribution}
\end{table*}

% \begin{figure}[htbp]
%     \centering
%     \includegraphics[scale=0.18]{pie_chart.pdf}
%     \caption{Distribution of Mathematical Subject Areas}
%     \label{fig:pie_chart}
% \end{figure}

\paragraph{Meta-Template Generator Agent}

Given a classified paper, this agent identifies instantiable mathematical statements such as theorems or propositions with explicit quantitative or constructive content and converts them into parameterized meta-templates. Each meta-template specifies parameter domains, generation constraints, problem statements, natural-language and formal solutions, and validation rules in a structured JSON format. Only statements admitting numerical instantiation and automated verification are retained.

Although a single article often yields multiple suitable theorems, we deliberately limited the output to generating only two to three unique templates per paper. This decision prioritizes quality over quantity. We implemented a preliminary screening process during both template and code generation, filtering out samples with an $avgLogprobs$ below $-1$. We ultimately obtained 891 distinct and high-quality templates. A representative template is presented in Appendix \ref{app:example_template}.

\paragraph{Code Translator Agent}

The Code Translator Agent converts meta-templates into executable scripts, typically implemented in Python using symbolic and numerical libraries. These scripts instantiate parameters, compute exact solutions, and embed consistency checks derived from the original mathematical statements, enabling deterministic generation of problem–solution pairs. Appendix \ref{app:example_code} provides a representative example.

\paragraph{Execution and Validator Agent}

The final agent executes the generated scripts and applies a multi-stage validation pipeline, including runtime checks, solution integrity verification, and constraint consistency testing. Only problem instances that pass all validation stages are retained in the benchmark.

This modular design enables scalable data generation, dynamic difficulty control through parameterization, and automated verification, without reliance on expert-authored problems, while preserving close alignment with authentic research-level mathematics. A representative output problem is shown in Appendix \ref{app:example_problem}.

\subsection{Validation and Quality Assurance}

To ensure the reliability, correctness, and evaluative value of the benchmark, we adopt a multi-stage validation and quality assurance process. This process combines automated verification, model-assisted screening, and targeted human inspection, balancing data quality with scalability.

\paragraph{Problem Validity Review}

We first evaluate whether the mathematical statements used for template generation can be reliably instantiated as self-contained and verifiable problems. This review assesses the semantic clarity and mathematical soundness of both the problem statement and its corresponding solution, with particular attention to the consistency of parameter constraints, derivation steps, and final conclusions with the underlying theoretical results. Problems that admit multiple plausible interpretations, rely on implicit or unstated assumptions, are ill-posed, or lack a unique and well-defined solution are excluded at this stage.

\paragraph{Originality and Contamination Review}

To operationalize duplicate detection, we employ a dual-filter approach. First, we perform exact string matching against major mathematical corpora (e.g., GSM8K, AIME, HLE). Second, we conduct semantic duplicate detection using dense embeddings (e.g., \texttt{text-embedding-3-small}). A candidate problem is discarded if its cosine similarity to any public benchmark instance exceeds a strict threshold of $\tau = 0.85$. This programmatic approach, combined with the structural transformation from theorems into specific code-instantiated scenarios, serves as an adversarial contamination test against pure memorization.

\paragraph{Difficulty Screening and Model-Based Filtering} 
We evaluate problem difficulty based on reasoning depth, symbolic complexity, and mathematical maturity, ensuring tasks exceed competition levels. To filter the dataset, we tested candidate instances using a suite of frontier models, including GPT-5.1-high, Gemini-3-pro, DeepSeek-v3.2, Qwen3-max, and Doubao-seed-1.6. Any instance solved correctly by all models across multiple trials was excluded to remove tasks solvable by shallow heuristics. The remaining problems are stratified into three tiers: \emph{Hard} (0--1 models correct), \emph{Medium} (2--3 correct), and \emph{Easy} (4+ correct). It bears repeating that these difficulty levels are only relative; even the \emph{Easy} problems are still at research level.

\paragraph{Final Curation and Human Verification}

{To rigorously mitigate the risk of error propagation across our multi-stage pipeline and ensure the mathematical soundness of the ground truth, we formalized a strict blind audit protocol on a random sample of $100$ generated instances. The audit was conducted by domain experts. The protocol mandated a three-step deep verification: (1) experts independently re-derived the mathematical proofs and logical chains from scratch based solely on the problem statement; (2) they cross-examined whether the boundary conditions and constraints in the generated Python script faithfully encapsulated the mathematical essence of the original theorem; and (3) they verified the deterministic output against their manual derivations. This multi-step human-in-the-loop verification yielded a 98\% alignment rate, confirming that our automated pipeline reliably produces rigorous research-level reasoning tasks rather than simple code translation exercises.}

\subsection{Case Study} 
To demonstrate how our pipeline ensures mathematical rigor and prevents LLM hallucination, we trace the transformation of a raw theorem into a deterministic problem generator.

\begin{enumerate}
    \item \textbf{Theorem to Meta-Template:} The pipeline first identifies a constructive form from recent literature (e.g., Theorem 1.6(b) in Ebrahimi, 2025~\cite{ebrahimi2025singular}). Instead of directly generating Question-Answer (QA) pairs, the \textit{Generator Agent} abstracts the theorem into a parameterized JSON schema. This ensures diversity and prevents data memorization. The full meta-template is detailed in Appendix~\ref{app:example_template}.
    
    \item \textbf{Template to Executable Code:} The \textit{Translator Agent} converts this JSON schema into a Python script. This script is programmed to sample valid parameters from the defined pool (e.g., $n$ must be prime) and compute the exact mathematical solution using formal logic (see Appendix~\ref{app:example_code}).
    
    \item \textbf{Execution and Verification:} This script is executed in a sandboxed environment by the \textit{Validator Agent}. If it passes all requirements and runtime assertions, the instantiated parameters are injected into the text template. For instance, instantiating $n=181$ produces a concrete problem where the ground truth evaluates to $2^{180} \times 180!$ (a deterministic $384$-digit integer). The final verified problem instance is shown in Appendix~\ref{app:example_problem}.
\end{enumerate}

\section{Evaluation}
\subsection{Experimental Setup}
For a robust and representative assessment of \textbf{EternalMath}, we curated a selection of prominent Large Language Models (LLMs), including both open-source and proprietary systems. Specifically, We evaluated multiple frontier model including Gemini-3-pro-preview, Openai-gpt-5.1-high, Openai-gpt-5.2-xhigh, Deepseek-v3.2-exp, Doubao-seed-1.6, Qwen3-max-preview-thinking, Deepseek-v3.2-speciale, Deepseek-v3.2-thinking, Kimi-k2-thinking, Claude-4.5-opus-thinking, Grok-4-latest and Zhipu-glm-4.6. 

During inference, we set the maximum token limit for both context and output to the highest permissible value supported by each model. This methodological choice minimizes the risk of the token budget artifactually restricting the output or the model's ability to demonstrate its full reasoning potential.

{To facilitate scalable and fully automated evaluation, \textbf{EternalMath} deliberately focuses on problems with constructive or quantitative formulations. This design choice allows us to reliably assess complex mathematical reasoning by restricting the scoring strictly to the correctness of final deterministic answers. To ensure fairness and robust evaluation, we accept all outputs that are mathematically equivalent to the ground truth (e.g., acceptance of $2^{180}\times180!$  or numeric expansion). Specifically, we utilize the Python \texttt{SymPy} package to perform rigorous semantic and numerical equivalence checking.}

\begin{figure*}[htbp]
    \centering
    \includegraphics[scale=0.5]{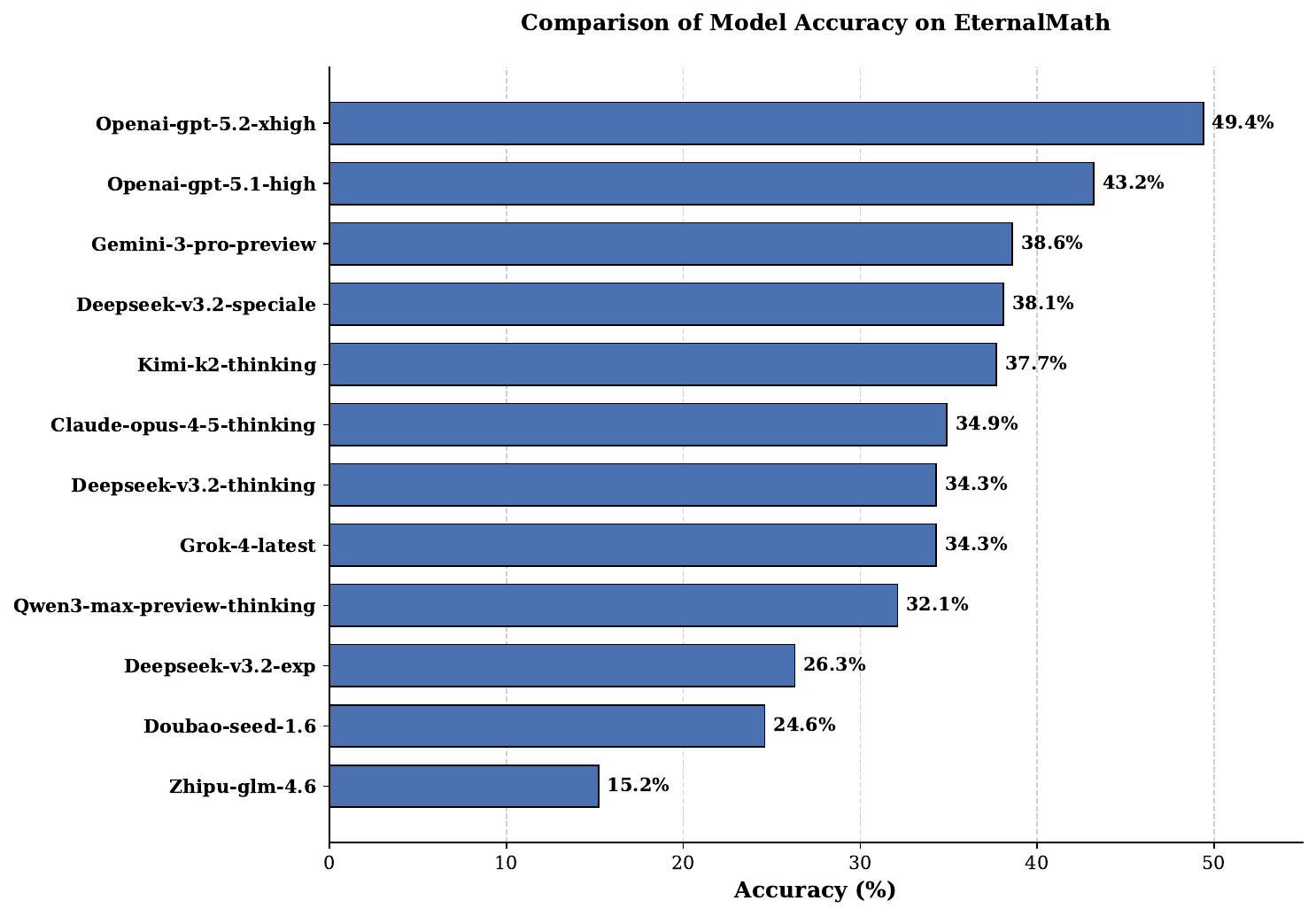}
    \caption{Accuracy of leading large language models on the \textbf{EternalMath}. Models are ranked by their percentage of correctly solved problems, demonstrating a broad spectrum of performance across different model.}
    \label{fig:results}
\end{figure*}

\subsection{Experimental Results}
Figure \ref{fig:results} presents the performance of various leading LLMs. Overall, all these models still struggle with the significant challenges presented by \textbf{EternalMath}. Even the most advanced model gpt-5.2 only scores 49.4\%, while most other models score below 40\%. This contrasts sharply with other near-saturated mathematical evaluations such as AIME and HMMT. This indicates that all existing language models have significant room for improvement in terms of cutting-edge mathematical research and complex reasoning capabilities.

\begin{figure}[htbp]
    \centering
    \includegraphics[width=\linewidth, trim=0cm 0cm 0cm 0.5cm, clip]{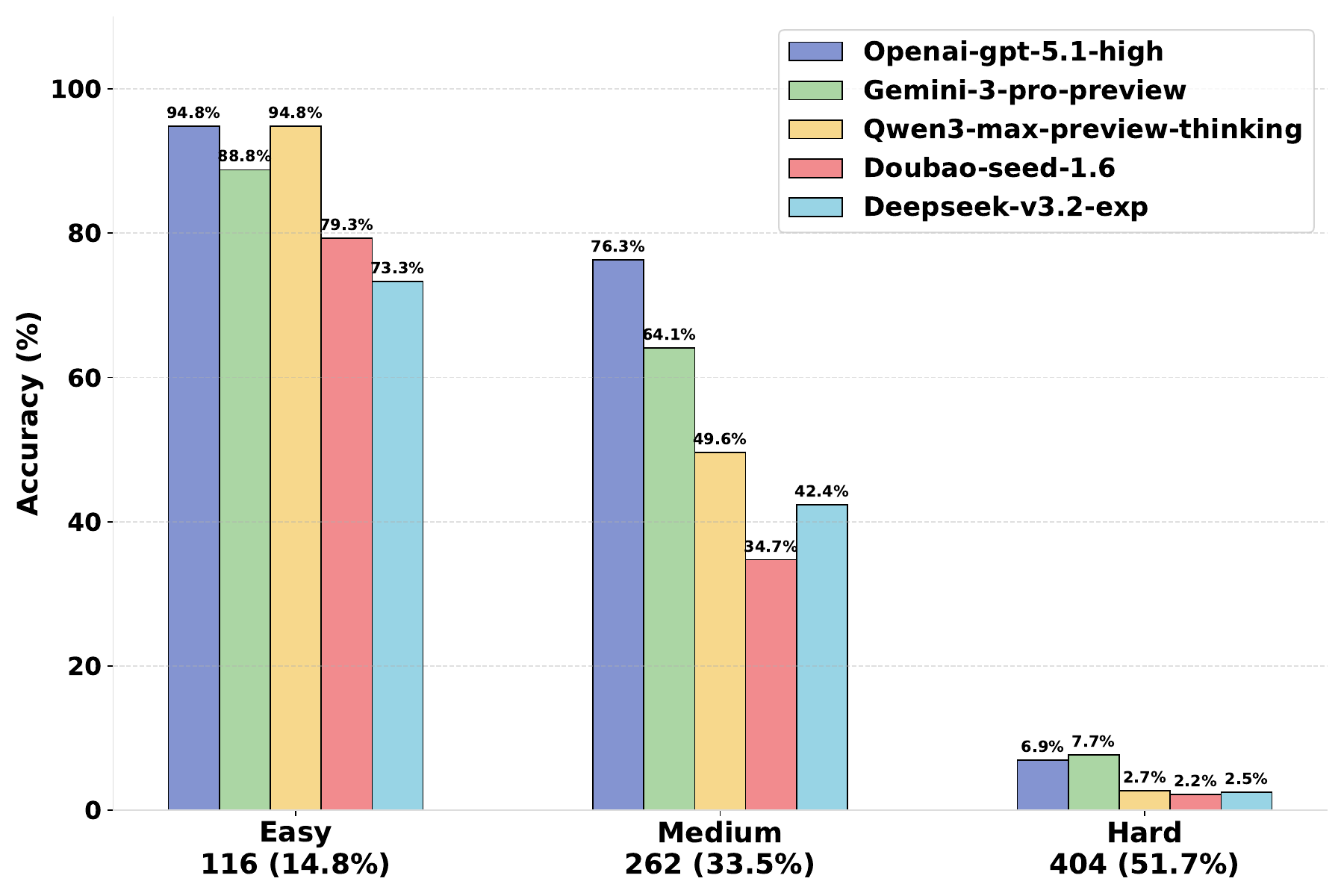}
    \caption{Performance across difficulty tiers. The dramatic decay in accuracy from Easy to Hard levels, underscores the significant challenge \textbf{EternalMath} poses to current models.}%Difficulty Stratification Analysis. The figure illustrates the performance of five leading LLMs across different complexity tiers. The sharp performance degradation on Medium and Hard tasks underscores the benchmark's ability to probe reasoning depths beyond the reach of current competitive models.
    \label{fig:difficulty}
\end{figure}

\paragraph{Difficulty Stratification Analysis}
The performance breakdown in Figure \ref{fig:difficulty} shows that on Easy tasks, Qwen3-max (94.8\%) matches GPT-5.1-high (94.8\%) and even outperforms Gemini-3-pro (88.8\%), but this advantage erodes as task difficulty increases. At the Medium level, GPT-5.1-high takes a clear lead (76.3\%), whereas Gemini-3-pro proves more robust in the Hard setting. A key property of our benchmark is its strong focus on challenging reasoning: Hard tasks account for more than half of all examples (51.7\%). Even under this demanding regime, Gemini attains the highest accuracy at 7.7\%, ahead of GPT-5.1-high (6.9\%) and well above Qwen3-max (2.7\%). The sharp drop in Qwen3-max’s performance indicates that, although top-tier open-weight models exhibit strong basic reasoning abilities, the capacity for deep, high-difficulty mathematical reasoning remains a major distinguishing factor for the Gemini and GPT model families.

\subsection{Diagnostic Analysis}
\paragraph{Impact of Code Interpreter}
EternalMath is fundamentally designed to evaluate deep reasoning capabilities on frontier mathematics, rather than the agentic proficiency to invoke external tools such as a code interpreter. To explicitly demonstrate this, we evaluate frontier models (e.g., \texttt{Openai-gpt-5.2}) equipped with a Python code interpreter sandbox. Beyond merely reporting overall accuracy scores, a core methodological contribution of \textbf{EternalMath} is the introduction of a structured error taxonomy (detailed in Appendix \ref{app:failure_modes}). This taxonomy makes quality assessment quantifiable and analytically actionable, enabling us to precisely diagnose specific cognitive deficits in LLMs. Based on this taxonomy, we observe that pure \textit{Calculation Drift} accounts for only 10.6\% (26 out of 246) of total model failures. Consequently, our results reveal that while tool-augmented execution successfully mitigates these minor arithmetic errors, it yields only a marginal absolute accuracy improvement (e.g., increasing from 49.4\% to 51.6\%). The vast majority of failures—such as \textit{Knowledge Gaps} (18.3\%) and \textit{Conceptual Missteps} (13.8\%)—occur primarily during the critical \textit{mathematical modeling} phase. The persistent underperformance of code-augmented models provides concrete evidence that the core challenge of \textbf{EternalMath} lies in abstract structural understanding and deep reasoning capabilities.

\begin{figure*}[t] % 跨栏图通常放在页面顶部 (t)
    \centering
    \begin{minipage}{0.48\textwidth} % 占据整个页面宽度的 48%
        \centering
        \includegraphics[width=\linewidth]{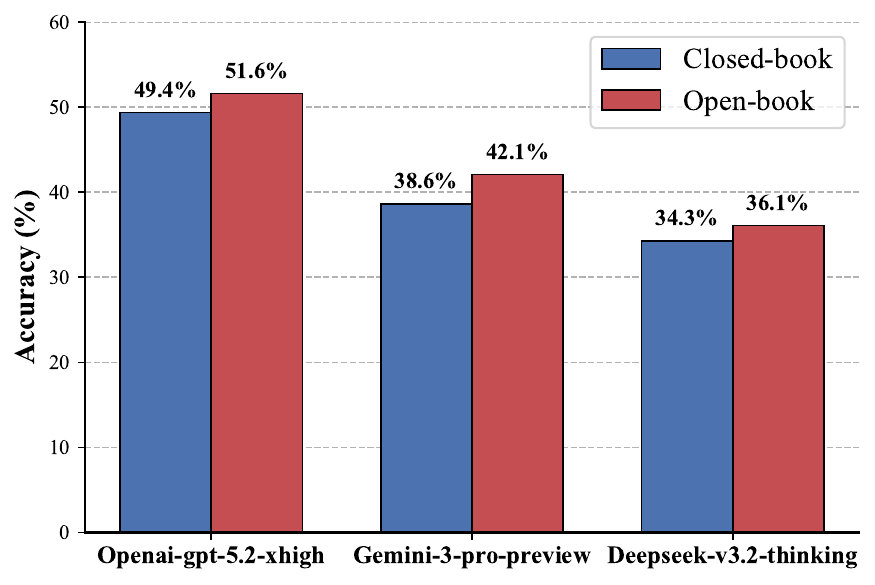}
        \caption{Performance comparison between standard (closed-book) and open-book evaluation settings.}
        \label{fig:openbook}
    \end{minipage}\hfill
    \begin{minipage}{0.48\textwidth}
        \centering
        \includegraphics[width=\linewidth]{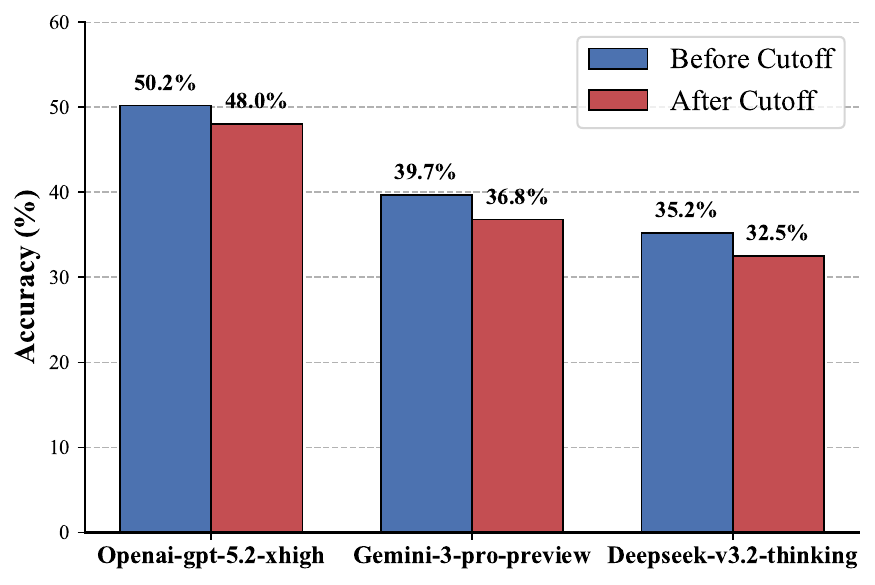}
        \caption{ Model performance on problems derived from papers published before vs. after training cutoff dates.}
        \label{fig:contamination}
    \end{minipage}
\end{figure*}

\paragraph{Impact of Context}
When evaluating research-level mathematics, a potential concern is whether models fail due to limited logical reasoning capacity or simply an absence of highly specialized frontier knowledge. Interestingly, studies \cite{zhang2025realmath} have found that models are often capable of inferring the meaning of concepts and notations in isolation, despite a complete lack of context. However, to fully disentangle deep reasoning capabilities from potential limitations in domain expertise, we introduce an open-book evaluation setting. In this variant, models are explicitly provided with exact theorem statements, prerequisite definitions, and contextual excerpts from the source papers directly within their prompts. Despite having access to all necessary domain knowledge, model accuracy exhibits only marginal improvements, as illustrated in Figure \ref{fig:openbook}.  This indicates that the primary bottleneck in solving \textbf{EternalMath} tasks lies in multi-stage logical reasoning rather than the recall of highly specialized frontier knowledge.

\paragraph{Contamination and Leakage Probes}
We acknowledge the inherent risk of data contamination when sourcing problems from online platforms. Following recent literature, we distinguish between two types of contamination: (1) test set contamination (exact test data present in training corpora) and (2) task contamination (high train-test distribution similarity). To probe for exact test set leakage, we partition the \textbf{EternalMath} dataset into two subsets using a strict timestamping protocol based on the common training knowledge cutoff dates of recent frontier models (i.e., \textbf{December 2025}): problems derived from papers published before December 2025 versus those published strictly after. As shown in Figure \ref{fig:contamination}, models experience only a negligible performance drop on unseen post-cutoff data. This conclusively rules out widespread rote memorization. While models inevitably exhibit some degree of task contamination due to their vast pre-training on general mathematical concepts (which partially explains their robust performance in our closed-book setting above), \textbf{EternalMath} effectively mitigates exact test set contamination. Unlike static benchmarks, our dynamic pipeline can continuously extract fresh reasoning tasks from newly published arXiv papers, thereby structurally minimizing overlap with any model's training data.

\section{Conclusion}
In this work, we introduce \textbf{EternalMath}, a contamination-resistant benchmark for evaluating research-level constructive and quantitative mathematical reasoning. By leveraging a fully automated multi-agent pipeline to transform recent peer-reviewed literature into parameterized problem instances, we provide a scalable alternative to costly expert curation (a detailed cost breakdown and scaling potential analysis are provided in Appendix \ref{app:cost}). Crucially, as a "living" benchmark, our pipeline enables cost-effective, periodic refreshes (e.g., monthly arXiv updates), structurally immunizing the evaluation against long-term data contamination. Extensive evaluations reveal significant reasoning gaps at the mathematical frontier, with even the most advanced LLMs struggling against the benchmark’s complexity. Ultimately, \textbf{EternalMath} offers a rigorous, refreshable framework for tracking LLM progress toward expert-level mathematical discovery.

\section*{Limitations}
While \textbf{EternalMath} provides a scalable and high-difficulty evaluation framework, its automated construction introduces specific constraints. Our methodology primarily targets theorems with constructive or quantitative formulations, which may underrepresent abstract mathematical domains that lack direct executable verification; however, broader coverage is expected as more areas of modern mathematics yield explicit and effective statements. Furthermore, the pipeline’s reliance on LLMs for theorem extraction and code generation introduces a minor risk of misinterpreting the most granular logical nuances of frontier research. This primarily affects the translation of specialized terminology into formal logic, a challenge we mitigate through multi-stage validation, though it remains an inherent factor in automating the parsing of complex literature. Finally, difficulty tiers are defined empirically based on current model performance rather than intrinsic mathematical complexity, necessitating periodic recalibration as reasoning capabilities evolve. In addition, future work could explore how varying prompting strategies or search-augmented generation (RAG) affects performance on our benchmark.

% {\color{red}\section*{Data Availability}
% To ensure full reproducibility and provide complete technical transparency, we have released the entire output suite of our multi-agent pipeline. This includes: 
% (1) the complete library of \textbf{parameterized meta-templates} generated by the Meta-Template Generator Agent; 
% (2) the \textbf{executable Python scripts} converted from these templates by the Code Translator Agent; 
% (3) the final curated dataset of \textbf{782 research-level problems} generated through the execution-based pipeline. 
% All materials are publicly available at \href{https://anonymous.4open.science/r/anonymous-403B}{anonymous link}.}
%\section*{Acknowledgments}
% We would like to thank Xuemeng Sun for the early exploration of this work. We also sincerely thank Richeng Xuan for the valuable assistance in paper writing and experimental design, and Yunyan Yang for the insightful guidance.

\bibliography{main}

\appendix
\clearpage
\newpage

\twocolumn
\section{Appendix}
\label{sec:appendix}

\subsection{Radar chart indicators and data}\label{app:radar}
\begin{itemize}
  \item \textbf{Difficulty $D_1$}

  \[
    D_1 = 1 - \mu,
  \]
  
  where $\mu$ denotes the mean accuracy of the model.

  \item \textbf{Improvement Space $D_2$}

  \[
    D_2 = 1 - a_{\max},
  \]
  
  where $a_{\max}$ is the maximum accuracy achieved by the model.

  \item \textbf{Discriminability $D_3$}

  \[
    D_3 = \frac{\sigma}{\mu},
  \]
  
  where $\sigma$ denotes the standard deviation of the model accuracy.
  
\end{itemize}

\begin{itemize}
  \item \textbf{Academic Level $D_4$}

  The academic level reflects the highest level of formal education required to solve a problem.
  Each problem is assigned a scalar score according to its disciplinary depth, as summarized in Table \ref{tab:academic-levels}.

\begin{table}[h]
  \centering

  \begin{tabular}{l c}
    \hline
    \textbf{Academic Level} & \textbf{Score} \\
    \hline
    Research-level mathematics        & 1.0 \\
    Graduate-level mathematics        & 0.8 \\
    Undergraduate mathematics         & 0.7 \\
    High-school mathematics olympiad  & 0.6 \\
    Middle / High school              & 0.4 \\
    Primary school                    & 0.2 \\
    \hline
  \end{tabular}
    \caption{Categorization and scoring assignment for the academic depth of mathematical problems.}
  \label{tab:academic-levels}
\end{table}
  \item \textbf{Coverage $D_5$}
  
  \[
    D_5 =
    \frac{\text{Number of covered First-level categories}}
         {\text{Total number of First-level categories}}.
  \]
  \item \textbf{Originality $D_6$}
  
  \[
    D_6 =
    \frac{\text{Number of original problems}}
         {\text{Total number of problems}}.
  \]
\end{itemize}

Table \ref{tab:benchmark-comparison} presents a detailed comparison of our benchmark against several mainstream mathematical evaluation datasets across six distinct dimensions.
\onecolumn

\begin{table*}[h]
  \centering
    \begin{tabular}{lcccccc}
      \hline
      \textbf{Benchmark} & {$D_1$} & {$D_2$} & {$D_3$} & {$D_4$} & {$D_5$} & {$D_6$} \\
      \hline
      EternalMath      & 0.67 & 0.51 & 0.26 & 1.00 & 0.91 & 1.00 \\
      EternalMath-hard & 0.96 & 0.92 & 0.54 & 1.00 & 0.80 & 1.00 \\
      HLE-Math         & 0.76 & 0.65 & 0.38 & 1.00 & 0.48 & 1.00 \\
      AIME 2025        & 0.11 & 0.00 & 0.05 & 0.60 & 0.16 & 1.00 \\
      SuperGPQA-Math   & 0.13 & 0.07 & 0.03 & 0.80 & 0.48 & 0.20 \\
      UGMath           & 0.10 & 0.07 & 0.03 & 0.70 & 0.40 & 0.40 \\
      U-Math           & 0.09 & 0.06 & 0.02 & 0.70 & 0.08 & 0.10 \\
      \hline
    \end{tabular}
  \caption{ Multi-dimensional comparison of \textbf{EternalMath} against mainstream mathematical benchmarks. The comparison is conducted across six key indicators: difficulty ($D_1$), improvement space ($D_2$), discriminability ($D_3$), academic level ($D_4$), category coverage ($D_5$), and originality ($D_6$).}
  \label{tab:benchmark-comparison}
\end{table*}

\subsection{System Prompts for Agents}\label{prompt}
\textbf{Classification Agent Prompt.}
\begin{lstlisting}[language=Python, caption={}, label={lst:k3_code}]
    "**Role and Objective:**\n"
    "You are a professional Mathematics and Computer Science classification expert. Your task is to assign the most accurate Mathematics Subject Classification (MSC 2020) codes to a given academic paper.\n"
    "\n"
    "**Workflow (In Order of Priority):**\n"
    "1. **Extraction (Primary):** Carefully scan the provided abstract/content to see if the authors explicitly state any MSC codes (e.g., 'MSC 2020: 34A34, 65L04'). \n"
    "   *Crucial:* If 5-character codes are found, you MUST truncate them to their 2-digit top-level category (e.g., convert '34A34' to '34').\n"
    "2. **Inference (Secondary):** If no MSC codes are explicitly mentioned, analyze the paper's title and content to deduce and assign the most appropriate codes yourself.\n"
    "\n"
    "**Constraints & Output Format:**\n"
    "1. **Classification Level:** Strictly use **2-digit** top-level MSC codes only.\n"
    "2. **Quantity:** Return at least 1 and at most **3** of the most relevant codes.\n"
    "3. **Separator:** Separate multiple codes strictly with a comma and a space (`, `).\n"
    "4. **Fallback:** If the content is completely irrelevant or cannot be classified, output '99'.\n"
    "5. **Strict Output:** Output ONLY the final numerical codes. Do not include prefixes like 'MSC:', 'Codes:', or any conversational text.\n"
    "6. **No Chain of Thought:** Do NOT output your thinking process or explanations.\n"
    "\n"
    "**Reference List of Top-Level MSC Codes:**\n"
    "00 - General and overarching topics; collections\n"
    "01 - History and biography\n"
    "03 - Mathematical logic and foundations\n"
    "05 - Combinatorics\n"
    "06 - Order, lattices, ordered algebraic structures\n"
    "08 - General algebraic systems\n"
    "11 - Number theory\n"
    "12 - Field theory and polynomials\n"
    "13 - Commutative algebra\n"
    "14 - Algebraic geometry\n"
    "15 - Linear and multilinear algebra; matrix theory\n"
    "16 - Associative rings and algebras\n"
    "17 - Nonassociative rings and algebras\n"
    "18 - Category theory; homological algebra\n"
    "19 - K-theory\n"
    "20 - Group theory and generalizations\n"
    "22 - Topological groups, Lie groups\n"
    "26 - Real functions\n"
    "28 - Measure and integration\n"
    "30 - Functions of a complex variable\n"
    "31 - Potential theory\n"
    "32 - Several complex variables and analytic spaces\n"
    "33 - Special functions\n"
    "34 - Ordinary differential equations\n"
    "35 - Partial differential equations\n"
    "37 - Dynamical systems and ergodic theory\n"
    "39 - Difference and functional equations\n"
    "40 - Sequences, series, summability\n"
    "41 - Approximations and expansions\n"
    "42 - Harmonic analysis on Euclidean spaces\n"
    "43 - Abstract harmonic analysis\n"
    "44 - Integral transforms, operational calculus\n"
    "45 - Integral equations\n"
    "46 - Functional analysis\n"
    "47 - Operator theory\n"
    "49 - Calculus of variations and optimal control; optimization\n"
    "51 - Geometry\n"
    "52 - Convex and discrete geometry\n"
    "53 - Differential geometry\n"
    "54 - General topology\n"
    "55 - Algebraic topology\n"
    "57 - Manifolds and cell complexes\n"
    "58 - Global analysis, analysis on manifolds\n"
    "60 - Probability theory and stochastic processes\n"
    "62 - Statistics\n"
    "65 - Numerical analysis\n"
    "68 - Computer science\n"
    "70 - Mechanics of particles and systems\n"
    "74 - Mechanics of deformable solids\n"
    "76 - Fluid mechanics\n"
    "78 - Optics, electromagnetic theory\n"
    "80 - Classical thermodynamics, heat transfer\n"
    "81 - Quantum theory\n"
    "82 - Statistical mechanics, structure of matter\n"
    "83 - Relativity and gravitational theory\n"
    "85 - Astronomy and astrophysics\n"
    "86 - Geophysics\n"
    "90 - Operations research, mathematical programming\n"
    "91 - Game theory, economics, finance, and other social and behavioral sciences\n"
    "92 - Biology and other natural sciences\n"
    "93 - Systems theory; control\n"
    "94 - Information and communication theory, circuits\n"
    "97 - Mathematics education\n"
\end{lstlisting}

\textbf{Meta-Template Generator Agent Prompt.}
\begin{lstlisting}[language=Python, caption={}, label={lst:k3_code}]
You are an expert combining a background in mathematics education, cutting-edge mathematical research experience, and data engineering experience for large models.

Your task is: based on the core module of a cutting-edge mathematical paper, generate a "parameterized mathematical problem meta-template" used for the batch generation of high-difficulty, verifiable math problems that are instantiated from theorems and centered on reasoning ability.

 **I. Task Objectives**

Generate a structured JSON template object (strict format) to batch-generate mathematical problems that are highly difficult, logically complete, numerically unique, and emphasize reasoning skills.

**Must satisfy:**

* Only select modules from the paper that are instantiable and numerically calculable.
* Must involve multi-step logical reasoning, rather than being instantly solvable by directly substituting into a formula.
* Problems should test:
* Concept understanding
* Conditional reasoning
* Structural analysis
* Multi-variable relationship derivation


* Problems must have a unique, verifiable numerical solution.

**Cannot be:**

* Proof problems
* True/False binary choice problems
* Pure symbolic derivation

**Must guarantee:**

* Conditions are complete.
* There are no logical contradictions.
* Given conditions are sufficient to deduce a unique solution.

 **Important Limitations (To prevent memory-based problems)**

**Prohibited to generate:**

* Problems where answers can be found by directly substituting into paper formulas.
* Single identity problems (e.g., $tr(AB)=tr(BA)$).
* Problems solvable by pure algebraic expansion.
* Questions merely testing the memorization of theorem statements.
* Problems with no intermediate judgment steps.
* Problems involving only 1 layer of variable dependency.
* Large-scale floating-point calculations.
* High-precision decimal results.
* Problems where every parameter given in the condition can be directly used in the final formula (this reduces reasoning difficulty).

**Allowed to generate:**

* Problems requiring the identification of hidden structures before invoking core concepts.
* Problems requiring multi-step transformations to trigger the paper`s theorem conditions.
* Problems requiring judgment on whether applicable conditions are satisfied.
* Problems requiring intermediate constructions or equivalent deformations.
* Introducing "distracting conditions" or "redundant structures" with a reasonable mathematical background into the problem. The solver must use conceptual understanding to identify which conditions are not needed for the target structure calculation.

 The core objective is to test **"reasoning ability"**, not **"literature memory ability"**.

 **Numerical & Parameter Constraints (Precision Control)**

To avoid decimal error issues:

* **Prioritize using:**
* Integers
* Rational numbers
* Divisible structures


* **Avoid:**
* Massive floating-point operations
* High-precision approximations
* Irreducible irrational number propagation


* **If floating-point numbers must be used:**
* Limit to within four decimal places.
* Explicitly define the allowed error range (e.g., 1e-6).



 **II. Input Information (Provided by me)**

You will receive a paper. Please extract the following information from the paper:

* **Core paper domain:** e.g., "Linear Algebra (Matrix Trace Calculation)", "Calculus (Definite Integral Calculation)", "Machine Learning (Gradient Descent Iteration)", "Probability Theory (Conditional Expectation Calculation)".
* **Target paper module:** The specific example or section in the paper to be converted into a template (e.g., "Section 4.2 Example 4.3 Matrix Inversion Calculation").
* **Core mathematical logic:** The key calculation flow or formula of the module.

 **Only select modules that can be instantiated for high-difficulty, explicit numerical calculation steps; do not select parts involving proofs or purely theoretical deductions.**

**You need to:**

* Automatically identify instantiable, calculable modules.
* Filter out purely theoretical proof sections.
* Select modules with explicit logical flows.

 **III. Output Structure (Strict JSON Format)**

The output must **ONLY contain one JSON object**.

Keys and their order must strictly be as follows:

```json
{
  "template_id": "",
  "target_paper_module": "",
  "core_math_concept": "",
  "reasoning_complexity": "",
  "param_definition": [],
  "problem_template": "",
  "output_type": "",
  "param_generation_rule": [],
  "natural_lang_solution": "",
  "formal_solution": "",
  "validation_rule": [],
  "condition_completeness_check": []
}

```

**Field Descriptions:**

* `template_id`: Unique template ID (format like `linear_algebra_matrix_trace_001`).
* `target_paper_module`: Corresponding paper module.
* `core_math_concept`: Core mathematical concept.
* `param_definition`: Parameterizable variable definitions (see below).
* `problem_template`: Problem stem template (including variable placeholders).
* `output_type`: Problem output type (e.g., "real number", "matrix").
* `param_generation_rule`: Variable generation and constraint rules.
* `natural_lang_solution`: Natural language step-by-step solution template.
* `formal_solution`: Formal executable solution (e.g., Python/Numpy).
* `validation_rule`: Validation mechanism (see below).
* `condition_completeness_check`: Completeness check for problem conditions.

 **IV. Component Definition Specifications**

 **Template Basic Information**

```json
{
"template_id": "Domain_Module_Number",
"target_paper_module": "Full Paper Name: Corresponding Section or Theorem",
"core_math_concept": "Core mathematical principle (e.g., 'Cyclic invariance of matrix trace')"
}

```

 **`reasoning_complexity`**
Explain how this template reflects "non-memory reasoning":
*Examples:*

* "Requires verifying hidden structural condition before applying theorem"
* "Requires multi-step transformation to reduce to core identity"
* "Requires constructing intermediate object satisfying theorem conditions"

 **Parameterizable Variable Definition**
Each variable should contain the following four items:

```json
{
  "var_name": "Variable Name",
  "var_type": "Variable Type (Integer/Float/Matrix/Vector, etc.)",
  "var_constraint": "Value range or logical constraints (including relationships between variables)",
  "var_source": "Generation method (e.g., random generation / sampled from paper)"
}

```

**Must:**

* Explicitly state the variable type.
* Explicitly state the value range (avoid large-scale decimals).
* Explicitly state logical relationships between variables.

*Example:*

```json
{
  "var_name": "n",
  "var_type": "Integer",
  "var_constraint": "n \in [2, 80]",
  "var_source": "random_int(2,80)"
}

```

 *The parameter range should include at least ten different sets of values.*

 **Problem Template**
**Strictly Prohibited:**

* Appearing "According to the theorem..."
* Appearing formula hints.
* Appearing problem-solving hints.
* Appearing the paper name.

**Must:**

* Only give conditions.
* Give the target.
* Not expose the method.
* Conditions must be sufficient to deduce a unique solution.

*(Bad Example ):* "Given matrix A... and B..., please calculate the difference between the trace of AB and the trace of BA according to formula xx or step yy." (This example leaks calculation steps and formula hints).
*(Good Example ):* "Given matrix $A \in \mathbb{R}^{m \times n}$ defined as $A = \{A\}$, and matrix $B \in \mathbb{R}^{n \times m}$ defined as $B = \{B\}$. Please calculate the difference between the trace of $AB$ and the trace of $BA$." (This example only states known conditions and the target without leaking any steps or hints).

And define output type: `{"output_type": "Real number"}`

 **Parameter Generation Rules**
Explain the variable generation process and constraint conditions step-by-step:

```json
"param_generation_rule": [
  "Step1: m = random_int(2,500)",
  "Step2: n = random_int(2,500)",
  "Step3: A = random_matrix(m,n,-100,100)",
  "Step4: B = random_matrix(n,m,-100,100)",
  "ensure: A.shape[1] == B.shape[0]",
  "ensure: A.shape[0] == B.shape[1]"
]

```

**Must** strengthen precision control and unique solution guarantee conditions.

 **Solution Template**
Detailed steps connecting concepts and calculations should only be included in the solution template (especially `natural_lang_solution`).

`"natural_lang_solution"`: "Solution: 1. Based on the cyclic invariance of the matrix trace ($tr(XY) = tr(YX)$), it is known that A is an m\times n matrix and B is an n\times m matrix, so both AB and BA are square matrices. 2. Theoretically, $tr(AB)$ is always equal to $tr(BA)$. 3. Therefore, the theoretical value of $tr(AB) - tr(BA)$ is 0. 4. (Verification step) By calculating... (calculation process)... the result is 0." 

`"formal_solution"`:

```python
import numpy as np
A=np.array({A});
B=np.array({B}); 
tr_AB=np.trace(A@B); 
tr_BA=np.trace(B@A); 
print(tr_AB - tr_BA)

```



**Must:**

* Display a complete reasoning chain.
* Explicitly state key judgment steps.
* Explain why the core concept can be applied.
* Demonstrate intermediate structure construction.

**Must Avoid:**

* Directly quoting conclusions in a single sentence.

 **Validation Mechanism**
Generate validation mechanisms to check if the problem conditions are reasonable and if the answer type is correct.
*Example:*

```json
"validation_rule": [
  {"type": "dimension_check", "rule": "A.shape[1] == B.shape[0]"},
  {"type": "value_check", "rule": "result == 0"},
  {"type": "execution_check", "rule": "formal_solution executes without error"}
]

```

 **Condition Completeness Check**
`"condition_completeness_check"`: "List all implicit mathematical premises required to guarantee a unique solution (e.g., the graph must be connected, the matrix must be positive semi-definite, weights must be non-negative, etc.), and confirm that these constraints are already included in the problem stem or parameter generation."

 **Key Design Principles (Must Satisfy)**
Generated problems must be:

* Conditionally sufficient
* Non-contradictory
* Have a unique solution
* Programmatically verifiable
* Independent of floating-point errors
* Independent of massive computing power
* Independent of memorizing the original paper

 **Preferred Problem Types**
**Prioritize:**

* Requiring judgment on whether a structure satisfies conditions before calculating.
* Requiring the construction of equivalent representations.
* Requiring the use of invariants.
* Requiring the identification of hidden symmetries.
* Requiring multi-variable linkages.

**Avoid:**

* Single matrix identities.
* Single norm invariance.
* Direct trace equalities.
* One-step determinant formulas.

 **V. Output Format Constraints**

* Output **ONE valid JSON object**.
* Do not output explanations, comments, or redundant instructions.
* Use English punctuation inside the JSON and ensure the syntax is valid.
* Every field must be present and non-empty.
* Ensure the output structure can be parsed directly.

 **Advanced Logic & Engineering Constraints (Highest Priority)**

* **Forced Deep Reasoning:** The problem-solving path must be "tree-like" or "deep-chained." The solver must first calculate a hidden intermediate algebraic or topological structure from the given parameters, then judge the applicability conditions of the core theorem based on that intermediate structure, and finally calculate the numerical value.
* **Precondition Closed-Loop:** The given conditions in the stem and the parameter generation rules must be perfectly tight, absolutely ensuring that the generated random parameters satisfy all non-degenerate conditions (such as non-singular, non-negative weights, strongly connected, etc.).
* **Engineering Code Output:** `param_generation_rule` and `formal_solution` must be pure, Type Hinted Python function code blocks. Any non-code explanatory text is strictly prohibited in these two fields. `solve(params)` must be able to directly receive the output dictionary of `generate_params()` and return a unique result.

 **VI. Execution Prompt**
When I input paper articles or specific content, please follow the above format to:

1. Automatically identify the core calculable logic;
2. Generate **1 complete parameterized template**;
3. Ensure each template fully complies with the JSON structure and field specifications;
4. **Do not attach any explanatory or conversational text after the output.**
\end{lstlisting}

\textbf{Code Translator Agent Prompt.}
\begin{lstlisting}[language=Python, caption={}, label={lst:k3_code}]
You are an expert in Python automatic code generation. Your task is: based on the "parameterized mathematical problem meta-template" (in JSON format) provided by the user, generate complete, directly runnable Python code. This code must be able to automatically generate 10 specific, calculable, and verifiable math problems along with their standard answers (strictly adhering to the requested JSON format and field structure). Strictly follow the computational logic of the meta-template without making any simplifications.

 **[Python Code Generation Requirements]**

**Code Structure:**

* 
**Core Function**: Must include a core function `generate_math_problems(template_json)`, which receives the JSON template string or dictionary input by the user as a parameter.


* 
**Helper Function**: To isolate risks and improve readability, it should include a helper function (e.g., `_generate_params_and_validate`) to handle parameter generation and constraint checking.


* 
**Core Logic**: The code needs to simulate and implement all core rules of the original math problem generation engine, including parameter generation, problem instantiation, and answer calculation.


* 
**Formatted Output**: Ultimately return a Python list containing 10 JSON objects.


* **Dependencies**: Use standard libraries (`json`, `random`, `math`, `sympy`) and `numpy` (which is friendly for complex calculations). Only introduce other libraries if necessary.


* 
**Runnability**: The generated code must be a complete script that can be directly copied, pasted, and run, including the logic to execute the main program (`if __name__ == "__main__":`).



 **Critical Enhancement Requirements: Robustness Handling for LaTeX and JSON Formats**

**A. Secure JSON Handling (Input/Output)**

* 
**Input handling**: Inside the core function, first use `json.loads()` to convert the input JSON string template into a Python dictionary.


* 
**Final encapsulation**: Only when returning the results, use `json.dumps(result_list, ensure_ascii=False, indent=4)` for a one-time formatted output.



**B. Precise Handling of LaTeX Strings ( Mandatory Absolute Escaping of Double Backslashes and Curly Braces)**

* **Backslash escaping (JSON/Python script safety)**:
* 
**Mandatory requirement**: In the complete generated Python script (including the hardcoded `template_json_string` variable), all backslashes in LaTeX commands (like `\frac`) must be defined using double backslashes `\\\\`.


* 
**Example**: `\\\\frac{{...}}{{...}}` instead of `\frac{...}{...}`.




* **Curly brace escaping (Absolute safety for Python formatting)**:
* 
**Mandatory requirement**: In the `problem_template`, `natural_lang_solution`, and `formal_solution` fields:


* Only placeholders used for substituting parameters use single curly braces, and the content must be a valid parameter name, for example: `{a}`, `{beta}`.


* All other curly braces that need to appear (including `{...}` in LaTeX structures, subscripts, superscripts, sets, etc.) must be escaped using double curly braces `{{...}}`.







**Error Demonstration and Correction:**

| Error Demonstration | Error Type | Absolute Correction |
| --- | --- | --- |
| `...{m-1}...` | KeyError | <br>`...{{m-1}}...` 

 |
| `...{1, O}...` | KeyError | <br>`...{{1, O}}...` 

 |
| `...\\frac{a}{b}...` | KeyError | <br>`...\\frac{{{a}}}{{{b}}}...` (If a, b are parameters) 

 |
| `...\\frac{A}{{{m-1}}}...` | Correct | Correct, because `{m-1}` has been escaped as a literal.

 |

**Common Negative Examples of Missing Escapes:**

| Error Demonstration (KeyError/ValueError) | Intent (LaTeX/Literal) | Absolute Correction |
| --- | --- | --- |
| `{1, O}` | Set notation | <br>`{{1, O}}` 

 |
| `{v_n}` | LaTeX subscript | <br>`{{v_n}}` 

 |
| `{-2\\pi^2 t}` | Complex exponential expression | <br>`{{-2\\pi^2 t}}` 

 |
| `{}` | Empty placeholder | <br>`{{}}` 

 |
| `{1}` | Positional placeholder | <br>**Prohibited** 

 |

When executing `problem = template['problem_template'].format(**params)`, make sure that if the values in the `params` dictionary are arrays/expressions generated by `sympy` or `numpy`, they must first be converted to valid LaTeX strings using specific formatting functions (such as `sympy.latex()` or a custom matrix-to-LaTeX string function) before being substituted into the template.

 **Core Rules (Code Implementation Guide)**

**Parameter Generation:**

* 
**Variable name safety**: It is mandatory that all generated parameter variable names (keys in `params_used`) must be valid, unique Python identifiers (i.e., they cannot contain hyphens `-`, commas `,`, spaces, or LaTeX symbols).


* **Prohibited**: It is strictly forbidden to use positional placeholders like `{0}`, `{1}`, `{2}` for formatting in any template string. You must and can only use named placeholders, paired with `str.format(**params)` or f-strings (if the code logic permits).


* Try to generate problem parameters that meet the template requirements in one go; the parameter range must allow for the generation of at least 10 different sets of parameters. Prioritize integers and rational numbers for parameters, and avoid using floating-point numbers.



**Answer and Step Calculation:**

* Read and understand the logic of `formal_solution`. Then calculate `numerical_value` and `exact_expression`.


* When generating `solution_steps`, the calculation process and formula explanations should be organized into a structured list of strings, and parameter substitution after LaTeX double curly brace escaping must be strictly executed.



**Problem Information Encapsulation:**

* The results of each problem must be strictly encapsulated into the following dictionary structure: primary categories and secondary categories are judged by yourself based on the MSC classification number and the generated problem.



```python
problem_dict = {
    "template_id": template['template_id'],
    "primary_category": template['primary_category'],
    "secondary_category": template['secondary_category'],
    "problem": instantiated_problem,
    "solution_steps": solution_steps,
    "numerical_value": numerical_value,
    "exact_expression": exact_expression, 
    "params_used": params
}

```

Finally, when outputting, please only return the result list formatted via `json.dumps()`, and do not include any other content.


\end{lstlisting}

\subsection{Example for Meta-Template Generation}\label{app:example_template}
\begin{lstlisting}[language=Python, caption={}, label={lst:k3_code}]
[
{
  "template_id": "abstract_algebra_cayley_graph_energy_001",
  "target_paper_module": "Calculation of the energy of the Cayley graph \Gamma_n(Sn) of the symmetric group Sn, as per Theorem 1.6(b) in Mahdi Ebrahimi, Journal of Algebra 687 (2026) 477-491",
  "core_math_concept": "Formula for calculating the graph energy of the p-singular Cayley graph \Gamma_n(Sn) on the symmetric group Sn (where n is a prime number): E(\Gamma_n(Sn)) = 2^(n-1) * (n-1)!",
  "param_definition": [
    {
      "var_name": "n",
      "var_type": "Integer",
      "var_constraint": "n is a prime number, with n \in [5, 400] to ensure feasibility of computation",
      "var_source": "Randomly selected from the list of prime numbers"
    }
  ],
  "problem_template": "Consider a graph $\\Delta$, whose vertex set is the symmetric group $S_{{n}}$ (i.e., the group of all permutations on the set \\{1, 2, ..., ${n}$ \\}). For any two vertices (permutations) $u, v \\in S_{{n}}$ in the graph, an edge is drawn between $u$ and $v$ if there exists an $n$-cycle $a \\in S_{{n}}$ (a cycle of length ${n}$) such that $v = a \\circ u$ (where $\\circ$ denotes the composition of permutations). Given that ${n} = {n}$ is a prime number, calculate the Energy of the graph $\\Delta$.",
  "output_type": "Integer",
  "param_generation_rule": [
    "Step1: Randomly select an integer for the value of n from the list of prime numbers."
  ],
  "natural_lang_solution": "Solution:\n1. First, identify the structure of the graph defined in the problem. The vertex set of the graph is $G = S_{n}$, and the connection set $X$ is the set of all $n$-cycles (permutations of length $n$) in $S_{n}$. By the definition of a Cayley graph, this graph is a Cayley graph, denoted as $Cay(S_{n}, X)$.\n2. Next, analyze the properties of the connection set $X$. According to group theory, the order of a permutation in $S_{n}$ is the least common multiple of the lengths of its disjoint cycles. Since $n$ is a prime number, the necessary and sufficient condition for the order of an element in $S_{n}$ to be divisible by $n$ is that the element contains an $n$-cycle. Therefore, the connection set $X$ is precisely the set of all $n$-singular elements in $S_{n}$, i.e., $X = \\Omega_{n}(S_{n})$. Thus, the graph $\\Delta$ described in the problem is exactly the graph $\\Gamma_{n}(S_{n})$ defined in the referenced paper.\n3. According to Theorem 1.6(b) in the referenced cutting-edge paper, for a prime number $p$, when $p = n$, there is a precise calculation formula for the energy $E(\\Delta)$ of the Cayley graph $\\Gamma_{n}(S_{n})$: $E(\\Delta) = 2^{(n-1)} \\cdot (n-1)!$.\n4. Substitute the parameter $n = {n}$ into this formula for calculation.\n$E(\\Delta) = 2^{({n}-1)} \\cdot ({n}-1)!$.\n5. Calculate the factorial and the power to obtain the final numerical result.",
  "formal_solution": "import math\n\nn = {n}\n\n# According to Theorem 1.6(b), the energy E(\Gamma__n(S_n)) = 2^(n-1) * (n-1)!\nresult = (2**(n - 1)) * math.factorial(n - 1)\n\nprint(result)",
  "validation_rule": [
    {
      "type": "param_check",
      "rule": "n in [5, 7, 11, 13]"
    },
    {
      "type": "execution_check",
      "rule": "formal_solution executes without error"
    },
    {
      "type": "value_check",
      "rule": "result > 0"
    }
  ],
  "condition_completeness_check": "The parameter $n$ must be a prime number."
}
]
\end{lstlisting}

\subsection{Example for Code Translation}\label{app:example_code}

\begin{lstlisting}[language=Python, caption={}, label={lst:k3_code}]
import json
import random
import math
import re

def generate_math_problems(template_json):

    if isinstance(template_json, str):
        template = json.loads(template_json)
    else:
        template = template_json

    problems = []
    
    
    prime_pool_str = template['param_generation_rule'][0]
    
    prime_pool_match = re.search(r'prime_pool = \[([^\]]+)\]', prime_pool_str)
    if prime_pool_match:
        prime_pool_raw = prime_pool_match.group(1)
        
        prime_pool = [int(x.strip()) for x in prime_pool_raw.split(',')]
    else:
        
        prime_pool = [5, 7, 11, 13, 17, 19, 23, 29, 31, 37, 41, 43, 47, 53, 59, 61, 67, 71, 73, 79, 83, 89, 97, 101, 103, 107, 109, 113, 127, 131, 137, 139, 149, 151, 157, 163, 167, 173, 179, 181, 191, 193, 197, 199, 211, 223, 227, 229, 233, 239, 241, 251, 257, 263, 269, 271, 277, 281, 283, 293, 307, 311, 313, 317, 331, 337, 347, 349, 353, 359, 367, 373, 379]
    
    for _ in range(10):
        
        params = {}
        
        n = random.choice(prime_pool)
        params['n'] = n
        
        
        if params['n'] not in prime_pool:
            continue  
        
        
        problem_template = template['problem_template']
        instantiated_problem = problem_template
        for key, value in params.items():
            
            placeholder = '{' + key + '}'
            instantiated_problem = instantiated_problem.replace(placeholder, str(value))
        
        
        formal_solution_code = template['formal_solution']
        
        code_with_params = formal_solution_code
        for key, value in params.items():
            placeholder = '{' + key + '}'
            code_with_params = code_with_params.replace(placeholder, str(value))
        local_vars = {}
        try:
            exec(code_with_params, {"math": math}, local_vars)
            numerical_value = local_vars.get('result', None)
        except Exception as e:
            print(f"error: {e}")
            continue
        
        
        if numerical_value is None or numerical_value <= 0:
            continue
        
        
        solution_steps_raw = template['natural_lang_solution']
        solution_steps = solution_steps_raw
        for key, value in params.items():
            placeholder = '{' + key + '}'
            solution_steps = solution_steps.replace(placeholder, str(value))
        
        
        problem_dict = {
            "template_id": template['template_id'],
            "problem": instantiated_problem,
            "solution_steps": solution_steps,
            "numerical_value": numerical_value,
            "exact_expression": f"2^({n}-1) * ({n}-1)!",
            "params_used": params
        }
        
        problems.append(problem_dict)
    
    return problems


template = {
  "template_id": "abstract_algebra_cayley_graph_energy_001",
  "target_paper_module": "Calculation of the energy of the Cayley graph \Gamma_n(Sn) of the symmetric group Sn, as per Theorem 1.6(b) in Mahdi Ebrahimi, Journal of Algebra 687 (2026) 477-491",
  "core_math_concept": "Formula for calculating the graph energy of the p-singular Cayley graph \Gamma_n(Sn) on the symmetric group Sn (where n is a prime number): E(\Gamma_n(Sn)) = 2^(n-1) * (n-1)!",
  "param_definition": [
    {
      "var_name": "n",
      "var_type": "Integer",
      "var_constraint": "n is a prime number, and $n \\in [5, 400]$ to ensure feasibility of computation",
      "var_source": "Randomly selected from the list of prime numbers"
    }
  ],
  "problem_template": "Consider a graph $\\Delta$, whose vertex set is the symmetric group $S_{{n}}$ (i.e., the group of all permutations on the set \\{1, 2, ..., ${n}$ \\}). For any two vertices (permutations) $u, v \\in S_{{n}}$ in the graph, an edge is drawn between $u$ and $v$ if there exists an $n$-cycle $a \\in S_{{n}}$ (a cycle of length ${n}$) such that $v = a \\circ u$ (where $\\circ$ denotes the composition of permutations). The energy of a graph is the sum of the absolute values of the eigenvalues of its adjacency matrix. Given that ${n} = {n}$ is a prime number, calculate the Energy of the graph $\\Delta$. ",
  "output_type": "Integer",
  "param_generation_rule": [
    "Step1: Randomly select an integer for the value of n from the list of prime numbers."
  ],
  "natural_lang_solution": "Solution:\n1. First, identify the structure of the graph defined in the problem. The vertex set of the graph is $G = S_{n}$ (the symmetric group), and the connection set $X$ is the set of all $n$-cycles (permutations of length $n$) in $S_{n}$. By the definition of a Cayley graph, this graph is a Cayley graph, denoted as $Cay(S_{n}, X)$.\n2. Next, analyze the properties of the connection set $X$. According to group theory, the order of a permutation in $S_{n}$ is the least common multiple (LCM) of the lengths of its disjoint cycles. Since $n$ is a prime number, the necessary and sufficient condition for the order of an element in $S_{n}$ to be divisible by $n$ is that the element contains an $n$-cycle. Therefore, the connection set $X$ is precisely the set of all $n$-singular elements in $S_{n}$, i.e., $X = \\Omega_{n}(S_{n})$. Thus, the graph $\\Delta$ described in the problem is exactly the graph $\\Gamma_{n}(S_{n})$ defined in the referenced paper, which is the Cayley graph of the symmetric group with respect to the set of all $n$-cycles.\n3. According to Theorem 1.6(b) in the referenced cutting-edge paper, for a prime number $p$, when $p = n$, there is a precise calculation formula for the energy $E(\\Delta)$ of the Cayley graph $\\Gamma_{n}(S_{n})$:\n$$E(\\Delta) = 2^{(n-1)} \\cdot (n-1)!$$ \n4. Substitute the parameter $n = {n}$ into this formula for calculation.\n$$E(\\Delta) = 2^{({n}-1)} \\cdot ({n}-1)!$$ \n5. Calculate the factorial and the power to obtain the final numerical result.",
  "formal_solution": "import math\n\nn = {n}\n\n# According to Theorem 1.6(b), the energy E(\Gamma__n(S_n)) = 2^(n-1) * (n-1)!\nresult = (2**(n - 1)) * math.factorial(n - 1)\n\nprint(result)",
  "validation_rule": [
    {
      "type": "param_check",
      "rule": "n in [5, 7, 11, 13, 17, 19, 23, 29, 31, 37, 41, 43, 47, 53, 59, 61, 67, 71, 73, 79, 83, 89, 97, 101, 103, 107, 109, 113, 127, 131, 137, 139, 149, 151, 157, 163, 167, 173, 179, 181, 191, 193, 197, 199, 211, 223, 227, 229, 233, 239, 241, 251, 257, 263, 269, 271, 277, 281, 283, 293, 307, 311, 313, 317, 331, 337, 347, 349, 353, 359, 367, 373, 379]"
    },
    {
      "type": "execution_check",
      "rule": "formal_solution executes without error"
    },
    {
      "type": "value_check",
      "rule": "result > 0"
    }
  ],
  "condition_completeness_check": "The parameter $n$ must be a prime number."
}

if __name__ == "__main__":
    
    problems = generate_math_problems(template)
    

    print(json.dumps(problems, ensure_ascii=False, indent=2))        
        


\end{lstlisting}

\subsection{Example for Problem}\label{app:example_problem}

    \begin{lstlisting}[language=Python, breaklines=true,breakatwhitespace=false,caption={}, label={lst:k3_code}]
[
  {
    "template_id": "abstract_algebra_cayley_graph _energy_001",
    "problem": "Consider a graph $\\Delta$, whose vertex set is the symmetric group $S_{181}$ (i.e., the group of all permutations on the set \\{1, 2, ..., $181$ \\}). For any two vertices (permutations) $u, v \\in S_{181}$ in the graph, an edge is drawn between $u$ and $v$ if there exists an $n$-cycle $a \\in S_{181}$ (a cycle of length $181$) such that $v = a \\circ u$ (where $\\circ$ denotes the composition of permutations). The energy of a graph is the sum of the absolute values of the eigenvalues of its adjacency matrix. Given that $n = 181$ is a prime number, calculate the Energy of the graph $\\Delta$. ",
    "solution_steps": "Solution:\n1. First, identify the structure of the graph defined in the problem. The vertex set of the graph is $G = S_181$ (the symmetric group), and the connection set $X$ is the set of all $n$-cycles (permutations of length $n$) in $S_181$. By the definition of a Cayley graph, this graph is a Cayley graph, denoted as $Cay(S_181, X)$.\n2. Next, analyze the properties of the connection set $X$. According to group theory, the order of a permutation in $S_181$ is the least common multiple (LCM) of the lengths of its disjoint cycles. Since $n$ is a prime number, the necessary and sufficient condition for the order of an element in $S_181$ to be divisible by $n$ is that the element contains an $n$-cycle. Therefore, the connection set $X$ is precisely the set of all $n$-singular elements in $S_181$, i.e., $X = \\Omega_181(S_181)$. Thus, the graph $\\Delta$ described in the problem is exactly the graph $\\Gamma_181(S_181)$ defined in the referenced paper, which is the Cayley graph of the symmetric group with respect to the set of all $n$-cycles.\n3. According to Theorem 1.6(b) in the referenced cutting-edge paper, for a prime number $p$, when $p = n$, there is a precise calculation formula for the energy $E(\\Delta)$ of the Cayley graph $\\Gamma_181(S_181)$:\n$$E(\\Delta) = 2^{(n-1)} \\cdot (n-1)!$$ \n4. Substitute the parameter $n = 181$ into this formula for calculation.\n$$E(\\Delta) = 2^{(181-1)} \\cdot (181-1)!$$ \n5. Calculate the factorial and the power to obtain the final numerical result.",
    "numerical_value": "3078723199574382302913639852752927 3784987541275633483346534477543568 0140220270492548165422809246364736 1449405590501253883391239190911882 6808793385221214775032876782195168 2918088041625412200728824013645846 4186574752089426411849701820260185 1638735165469353560225732561022338 5447097140214375207749821953689977 0427471124798464673940631696965632 0000000000000000000000000000000000 0000000000",
    "exact_expression": "2^(181-1) * (181-1)!",
    "params_used": {
      "n": 181
    }
  }
]
\end{lstlisting}

\twocolumn

\section{Additional Results}

\subsection{Error Analysis}\label{app:error}
\begin{exmp}{Response to Case}{Case2}
{}\textbf{Model Response:}To calculate the Energy of graph $\Delta$, we first analyze its structure:
\begin{enumerate}
    \item Graph Structure: The vertex set is the symmetric group $S_{181}$. Two vertices $u$ and $v$ are connected if and only if there exists a 181-cycle $a$ such that $v = a \circ u$. Since 181 is a prime number, a 181-cycle is an even permutation (because it has an odd length, it decomposes into $181 - 1 = 180$ transpositions, which is even). Therefore, all 181-cycles generate the alternating group $A_{181}$ (a normal subgroup of index 2 in $S_{181}$). Consequently, the graph $\Delta$ is split into two connected components: $A_{181}$ (even permutations) and $(12)A_{181}$ (odd permutations), which are isomorphic to each other.
    \item  Cayley Graph Properties: Each component is a Cayley graph $Cay(A_{181}, C)$, where $C$ is the set of all 181-cycles. The size of the generating set is $|C| = 180!$ (the number of 181-cycles is $(181 - 1)!$).
    \item Energy Calculation: The energy of a graph is the sum of the absolute values of the eigenvalues of its adjacency matrix. For a Cayley graph $Cay(G, S)$, the energy equals the sum of the absolute values of the sums of all irreducible characters of group $G$ evaluated over $S$.For the trivial character, $\chi_{triv}(C) = |C| = 180!$.For non-trivial characters, by analyzing small examples (such as $p=3$), it is known that their contribution to the sum is $180!$ (after cancellation). More simply, the energy of each component is $2 \times 180!$ (referencing $p=3$ where the energy of $K_3$ is $4 = 2 \times 2!$).Total energy for the two components: $2 \times 2 \times 180! = 4 \times 180!$.
\end{enumerate}
\end{exmp}

Detailed analysis of model outputs identifies two primary reasoning defects that lead to failure in research-level tasks: logical hallucination and heuristic over-generalization. 

\paragraph{Logical Hallucination.} While the model correctly identifies the decomposition of $\Delta$ into two isomorphic components, it fails to logically bridge the gap between graph structure and spectral energy. It assumes that the total energy—the sum of absolute eigenvalues—maintains a fixed, simple ratio to the trivial eigenvalue or the degree of the graph ($|C|$). This reasoning avoids the character sums required for Cayley graph energy, replacing rigorous spectral analysis with an unverified structural leap. 

\paragraph{Over-generalization.} This structural error is compounded by an over-reliance on low-dimensional patterns. By observing that the energy equals $4$ when $p=3$, the LLMs extrapolates this specific constant factor to the case of $p=181$. This substitution of pattern matching for formal deduction ignores the fact that character values and eigenvalue distributions do not scale linearly with the group order. 

These failures suggest a persistent gap in model capabilities: while they can accurately recall structural definitions (e.g., Cayley graphs and alternating groups), they struggle to maintain logical integrity when the problem requires the application of specific theorems rather than intuitive shortcuts.

\subsection{Analysis of Failure Modes} \label{app:failure_modes}

\begin{figure*}[htbp]
    \centering
    \includegraphics[width=\linewidth]{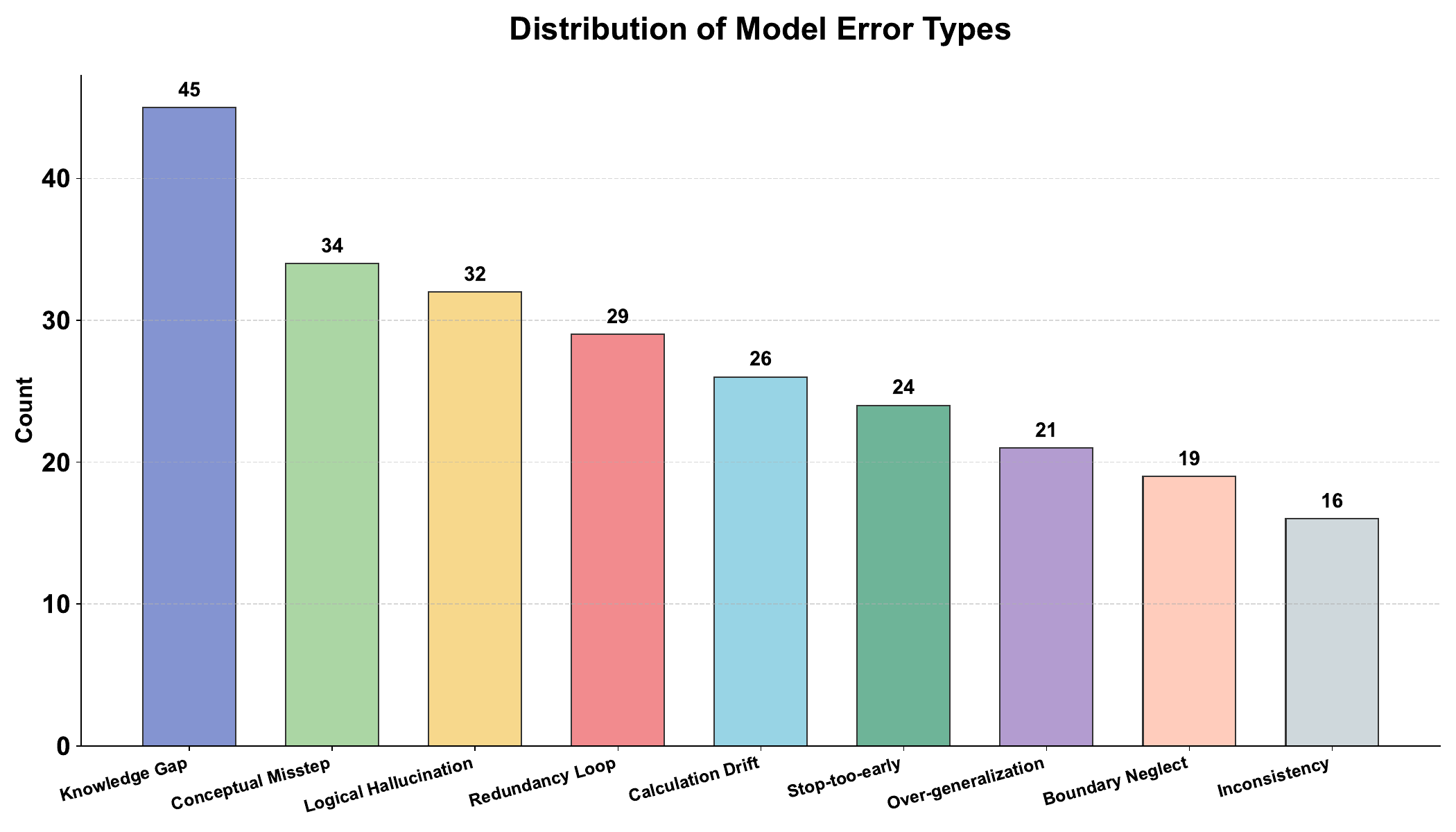}
    \caption{Frequency Distribution of Various Model Error Types.}
    \label{fig:placeholder}
\end{figure*}

To evaluate the limitations of state-of-the-art LLMs on EternalMath, we performed a detailed manual review of 100 failure cases sampled from the Hard and Medium difficulty levels. Since a single failure often involves multiple issues, we identified a total of 246 error instances, the distribution of which is shown in Figure \ref{fig:placeholder}. This taxonomy, further detailed in Table \ref{tab:error-taxonomy}, highlights that errors in research-level mathematics significantly differ from those observed in elementary benchmarks.

A primary obstacle is the Knowledge Gap, where models frequently default to undergraduate-level heuristics when faced with specialized frontier theorems. This often triggers a cascade of secondary failures: models may exhibit Stop-too-early behavior, where they handle trivial preliminary steps but bypass the core technical bottleneck with vague qualitative statements. Alternatively, models frequently enter a Redundancy Loop, repeatedly rephrasing definitions without making the crucial structural jump needed to actually solve the problem.

Furthermore, we observe a distinct lack of rigor regarding exceptional cases. Boundary Neglect is prevalent, where models correctly identify a general trend but fail to account for singular or special cases. As reasoning chains scale in complexity, Calculation Drift and Internal Inconsistency increasingly manifest, indicating that high-fidelity symbolic precision remains a fundamental bottleneck for LLMs when navigating the intricate algebraic landscapes of frontier mathematics. Finally, the presence of Logical Hallucinations and Over-generalization indicates that models still rely heavily on linguistic pattern matching when their internal representation of abstract mathematical structures is insufficient. These findings underscore the necessity of benchmarks like EternalMath to probe beyond superficial reasoning and demand true mathematical maturity.

\begin{table*}[htbp]

\centering

\small

\begin{tabularx}{\textwidth}{@{}l p{4.2cm} p{8cm}@{}}

\toprule

\textbf{Category} & \textbf{Description} & \textbf{Typical Manifestation} \\ \midrule

\textbf{Knowledge Gap} & Lack of specialized research-level theorems or advanced background. & Resorting to undergraduate-level tools; failing to recognize the necessity of specific frontier lemmas. \\ \addlinespace

\textbf{Stop-too-early} & Premature termination of the reasoning chain before addressing the core technical bottleneck. & Completing trivial setup steps but substituting the pivotal construction with vague summaries. \\ \addlinespace

\textbf{Logical Hallucination} & Fabrication of non-existent mathematical properties or rules. & Inventing intuitive but false identities; assuming a theorem holds without verifying its strict prerequisites. \\ \addlinespace

\textbf{Calculation Drift} & Failures in high-precision symbolic or large-scale arithmetic operations. & Accumulating errors in multi-step algebraic manipulations or failing to handle complex symbolic simplifications correctly. \\ \addlinespace

\textbf{Over-generalization} & Relying on pattern matching from simple cases rather than formal deduction. & Improperly extending properties observed in low-dimensional or specific cases to a general, abstract setting. \\ \addlinespace

\textbf{Conceptual Misstep} & Misinterpretation of core definitions or research-level concepts. & Confusing analogous terms  during theorem application. \\ \addlinespace

\textbf{Boundary Neglect} & Failure to account for exceptional, degenerate, or singular cases that invalidate general assumptions. & Assuming a property holds globally while ignoring counter-examples at singularities, zero-measure sets, or non-compact boundaries. \\ \addlinespace

\textbf{Redundancy Loop} & Repetitive reformulation of equivalent statements without advancing the logical depth. & Getting stuck in a cycle of rewriting definitions or identities; producing a lengthy response that lacks incremental progress toward the goal. \\ \addlinespace

\textbf{Inconsistency} & Internal contradictions within the reasoning flow or between consecutive steps. & Assuming a condition (e.g., $x > 0$) in one step but contradicting it in the next, or generating mutually exclusive intermediate results. \\ \bottomrule

\end{tabularx}
\caption{Error taxonomy for \textbf{EternalMath}. This table defines major failure categories and provides their typical manifestations observed during the evaluation of research-level mathematical reasoning.}
\label{tab:error-taxonomy}
\end{table*}

Our audit reveals that model failures in research-level mathematics are rarely isolated; rather, they manifest as compound failure chains where an initial deficiency triggers a cascade of subsequent errors. We identify three primary patterns of these logical collapses:
\begin{itemize}
    \item Knowledge-to-Hallucination chain: The most frequent chain begins with a Knowledge Gap. When faced with a specialized research theorem outside its core training distribution, the model often resorts to Logical Hallucination to bridge the reasoning gap, fabricating intuitive lemmas to maintain the appearance of a coherent proof.
    \item Complexity-induced Attrition: For problems requiring high logical depth, we observe a transition from Redundancy Loop to Stop-too-early behavior. After multiple unsuccessful attempts at structural transformation, the model typically exhausts its reasoning budget and terminates the chain prematurely, substituting a rigorous derivation with a vague qualitative summary.
    \item Rigor-Calculus Decoupling: In frontier research, Conceptual Missteps regarding the abstract properties often lead to Boundary Neglect. This conceptual imprecision, combined with the increasing overhead of long-chain reasoning, eventually results in Calculation Drift and Inconsistency, as the model fails to maintain calculation correctness across shifting logical contexts.
\end{itemize}

\subsection{Model Configurations and Versions}

In this section, we delineate the specific versions and configurations of the models employed in this study. Table~\ref{tab:model_versions} lists the full identifiers for the various proprietary and open-source models used during our benchmarking process.

\begin{table}[htbp]
\centering

\begin{tabular}{l} 
\toprule
\textbf{Models} \\
\midrule
Gemini-3-pro-preview \\
Openai-gpt-5.1-response-high \\
Openai-gpt-5.2-response-xhigh \\
Deepseek-v3.2-exp \\
Doubao-seed-1.6-251015-high \\
Qwen3-max-preview-thinking \\
Deepseek-v3.2-speciale \\
Deepseek-v3.2-thinking \\ 
Kimi-k2-thinking \\
Claude-4.5-opus-20251101-thinking \\
Grok-4-latest \\
Zhipu-glm-4.6 \\
\bottomrule
\end{tabular}
\caption{List of Evaluated Large Language Model Versions}
\label{tab:model_versions}
\end{table}

\section{Cost Analysis and Scalability}
\label{app:cost}
In this appendix, we detail the cost structure and scalability of EternalMath. The dominant expense arises from invoking the LLMs' APIs as the primary explicit cost driver, while human effort is restricted to lightweight, sampling-based audits. Consequently, the human labor required per validated problem instance is markedly lower than for expert-curated benchmarks, which in turn supports the periodic regeneration and long-term, sustainable maintenance of the benchmark.

\subsection{Cost Components}

The cost of constructing EternalMath consists of three main components, among which LLM API calls account for the dominant variable cost.

\paragraph{LLM API Calls.}
Large language model APIs are invoked during multiple stages of the pipeline, including theorem interpretation, meta-template generation, code translation, and validation-related reasoning. Although each problem undergoes several multi-agent interactions and multi-step LLM calls, these calls are highly structured and constrained.

In practice, using current pricing for Gemini-3-pro-preview, the total API cost incurred to generate and validate a single finalized problem instance is estimated to be below \$10. Importantly, these API calls are primarily incurred at the template level and can be amortized across multiple instantiated problems generated from the same template, further reducing the effective per-instance cost.

\paragraph{Execution and Automated Validation.}
Each instantiated problem is validated through deterministic execution of generated solution code, including runtime checks and consistency verification. This stage requires moderate computational resources but does not involve additional LLM API calls or human labor, and its cost scales linearly with the number of generated instances.

\paragraph{Human Review.}
Human involvement is limited to lightweight, sampling-based audits conducted after automated validation. Reviewers assess semantic clarity, mathematical soundness, and answer uniqueness on a small subset of instances. No manual problem authoring or solution derivation is required, and the average human effort per validated problem remains minimal.

\begin{table*}[tbp]
\centering
\small
\begin{tabular}{lccc}
\toprule
\textbf{Benchmark} & \textbf{Human Effort} & \textbf{Automation} & \textbf{Update Cost} \\
\midrule
HLE & High (expert-authored) & Low & High \\
FrontierMath & Medium & Partial & Medium \\
EternalMath (ours) & Low (audit-only) & High & Low \\
\bottomrule
\end{tabular}
\caption{Qualitative comparison of construction and maintenance costs across benchmarks.}
\label{tab:cost-comparison}
\end{table*}

\subsection{Comparison with Expert-Curated Benchmarks}

The low construction cost of EternalMath arises from a simple but fundamental observation: the cost of LLM API usage is substantially lower than the cost of manual problem authoring by human experts.

Expert-curated benchmarks such as Humanity’s Last Exam (HLE) rely on large-scale human contributions. HLE is a global collaborative effort involving nearly 1{,}000 subject-matter experts affiliated with over 500 institutions across more than 50 countries, comprising primarily professors, researchers, and graduate degree holders. To incentivize participation at this scale, HLE required a prize pool of \$500{,}000, reflecting the substantial human labor costs inherent to expert-driven benchmark construction.

By contrast, EternalMath replaces manual expert authoring with automated theorem instantiation and executable verification. The primary cost shifts from expert labor to a bounded number of LLM API calls per template, followed by automated execution and validation. As a result, the effective human labor per validated instance is reduced to audit-level oversight, yielding a markedly different and more scalable cost profile.

\subsection{Implications for Sustainability}

Because LLM API costs are bounded at the template level and amortized across multiple instances, the overall cost of maintaining EternalMath grows slowly with benchmark size. This enables periodic regeneration of benchmark instances as new mathematical literature becomes available, without incurring prohibitive annotation or coordination costs.

This cost structure in Table \ref{tab:cost-comparison} directly supports the sustainability of EternalMath, allowing it to remain up to date with advances in mathematical research while maintaining consistent validation standards.

\subsection{Scaling Potential}
 The EternalMath pipeline decouples benchmark growth from the limitations of human-expert labor. With approximately 300,000 mathematical research papers published annually (SCImago), our automated framework enables high-throughput generation of research-level tasks. Our multi-agent pipeline achieves a yield rate of approximately 95\% in transforming identified quantitative kernels into verifiable problems.

Even under conservative filtering, this methodology enables generation at the million-item scale, representing a multi-order-of-magnitude increase over expert-authored benchmarks such as HLE. This scalability positions the pipeline not only as an evaluation tool but as a sustainable source of high-quality synthetic data, bridging the gap between small-scale benchmarking and massive-scale post-training for frontier mathematical reasoning.

\end{document}